\pdfoutput=1

\documentclass[11pt]{article}

\usepackage[]{EMNLP2023}

\usepackage[T1]{fontenc}

\usepackage[utf8]{inputenc}

\usepackage{microtype}

\usepackage{inconsolata}
\usepackage{times}
\usepackage{latexsym}
\usepackage{amsfonts}
\usepackage{amsmath}
\usepackage{amssymb}
\usepackage{multicol}
\usepackage{multirow}
\usepackage{xspace}
\usepackage{booktabs}
\usepackage{bbding}
\usepackage{array}
\usepackage{threeparttable}
\usepackage{tcolorbox}
\usepackage{tabularx}
\usepackage{enumitem}
\usepackage[linesnumbered,ruled,vlined]{algorithm2e}
\usepackage{xcolor,colortbl}
\usepackage{setspace}
\usepackage{makecell}
\usepackage{listings}
\usepackage{xltabular}

\usepackage{cleveref}
\crefformat{section}{\S#2#1#3}
\crefformat{subsection}{\S#2#1#3}
\crefformat{subsubsection}{\S#2#1#3}
\crefrangeformat{section}{\S\S#3#1#4 to~#5#2#6}
\crefmultiformat{section}{\S\S#2#1#3}{ and~#2#1#3}{, #2#1#3}{ and~#2#1#3}
\crefmultiformat{subsection}{\S\S#2#1#3}{ and~#2#1#3}{, #2#1#3}{ and~#2#1#3}
\Crefformat{figure}{#2Fig.~#1#3}
\Crefmultiformat{figure}{Figs.~#2#1#3}{ and~#2#1#3}{, #2#1#3}{ and~#2#1#3}
\Crefformat{table}{#2Tab.~#1#3}
\Crefmultiformat{table}{Tabs.~#2#1#3}{ and~#2#1#3}{, #2#1#3}{ and~#2#1#3}
\Crefformat{appendix}{Appx.~\S#2#1#3}
\crefmultiformat{appendix}{Appx.~\S#2#1#3}{ and~#2#1#3}{, #2#1#3}{ and~#2#1#3}
\crefformat{algorithm}{Alg.~#2#1#3}
\Crefformat{equation}{Eq.~#2#1#3}

\newcommand{\xhdr}[1]{\vspace{0.3em}\noindent{{\bf #1.}}}
\newcommand{\modelname}{\textsc{Pivoine}\xspace}
\newcommand{\modelnamewex}{\textsc{P\underline{i}v\underline{oi}n\underline{e}}\xspace}

\newcommand{\modelnameex}{(\underline{I}nstruction-following \underline{O}pen-world \underline{I}nformation \underline{E}xtraction)}
\newcommand{\datasetname}{\textsc{InstructOpenWiki}\xspace}

\newcommand{\PartitionEx}{The partition ``Before 05/30/2022'' denotes mentions linked to the training ontology (seen entities), while ``After 05/30/2022'' denotes mentions linked to entities that are out of training ontology (unseen entities).\xspace}
\newcommand{\DevEx}{Subscript scores are the deviation under three different rephrased instructions for each sample.\xspace}

\title{\modelname: Instruction Tuning for Open-world Information Extraction}

\author{
Keming Lu$^{\dag}$\thanks{$^*$Work done during Keming Lu's internship at Tencent AI Lab.}, Xiaoman Pan$^{\ddag}$, Kaiqiang Song$^\ddag$, Hongming Zhang$^\ddag$, Dong Yu$^\ddag$, Jianshu Chen$^\ddag$
\\
$^\dag$University of Southern California, Los Angeles, CA\\
$^\ddag$Tencent AI Lab, Bellevue, WA \\ 
$^\dag$\texttt{keminglu@usc.edu}\\
$^\ddag$\texttt{\{xiaomanpan,riversong,hongmzhang,dyu,jianshuchen\}@global.tencent.com}\\
}

\begin{document}
\maketitle

\begin{abstract}
We consider the problem of Open-world Information Extraction~(Open-world IE), which extracts comprehensive entity profiles from unstructured texts. Different from the conventional closed-world setting of Information Extraction~(IE), Open-world IE considers a more general situation where entities and relations could be beyond a predefined ontology. More importantly, we seek to develop a large language model (LLM) that is able to perform Open-world IE to extract desirable entity profiles characterized by (possibly fine-grained) natural language instructions. We achieve this by finetuning LLMs using instruction tuning. In particular, we construct \datasetname, a substantial instruction tuning dataset for Open-world IE enriched with a comprehensive corpus, extensive annotations, and diverse instructions. We finetune the pretrained BLOOM models on \datasetname and obtain \modelname, an LLM for Open-world IE with strong instruction-following capabilities. Our experiments demonstrate that \modelname significantly outperforms traditional closed-world methods and other LLM baselines, displaying impressive generalization capabilities on both unseen instructions and out-of-ontology cases. 
Consequently, \modelname emerges as a promising solution to tackle the open-world challenge in IE effectively.\footnote{Checkpoints and datasets are available at \url{https://github.com/Lukeming-tsinghua/Instruction-Tuning-for-Open-world-IE}}
\end{abstract}

\section{Introduction}
Information extraction~(IE) aims to discern meaningful information from unstructured data sources~\cite{grishman2015information}.
A traditional IE pipeline contains an array of tasks, which include, but are not limited to, Named Entity Recognition~(NER)~\cite{lample-etal-2016-neural}, Entity Linking~(EL)~\cite{kolitsas-etal-2018-end}, Entity Typing~(ET)~\cite{ren-etal-2016-afet}, Relation Extraction~(RE)~\cite{huguet-cabot-navigli-2021-rebel-relation}, etc.
IE plays a vital role in knowledge graph construction~\cite{schneider2022decade}, search engine~\cite{wang2022webformer}, and document analysis~\cite{chiticariu2010enterprise,wang2018clinical,zhong2020does}.

\begin{figure}[t]
    \centering
    \includegraphics[width=\linewidth]{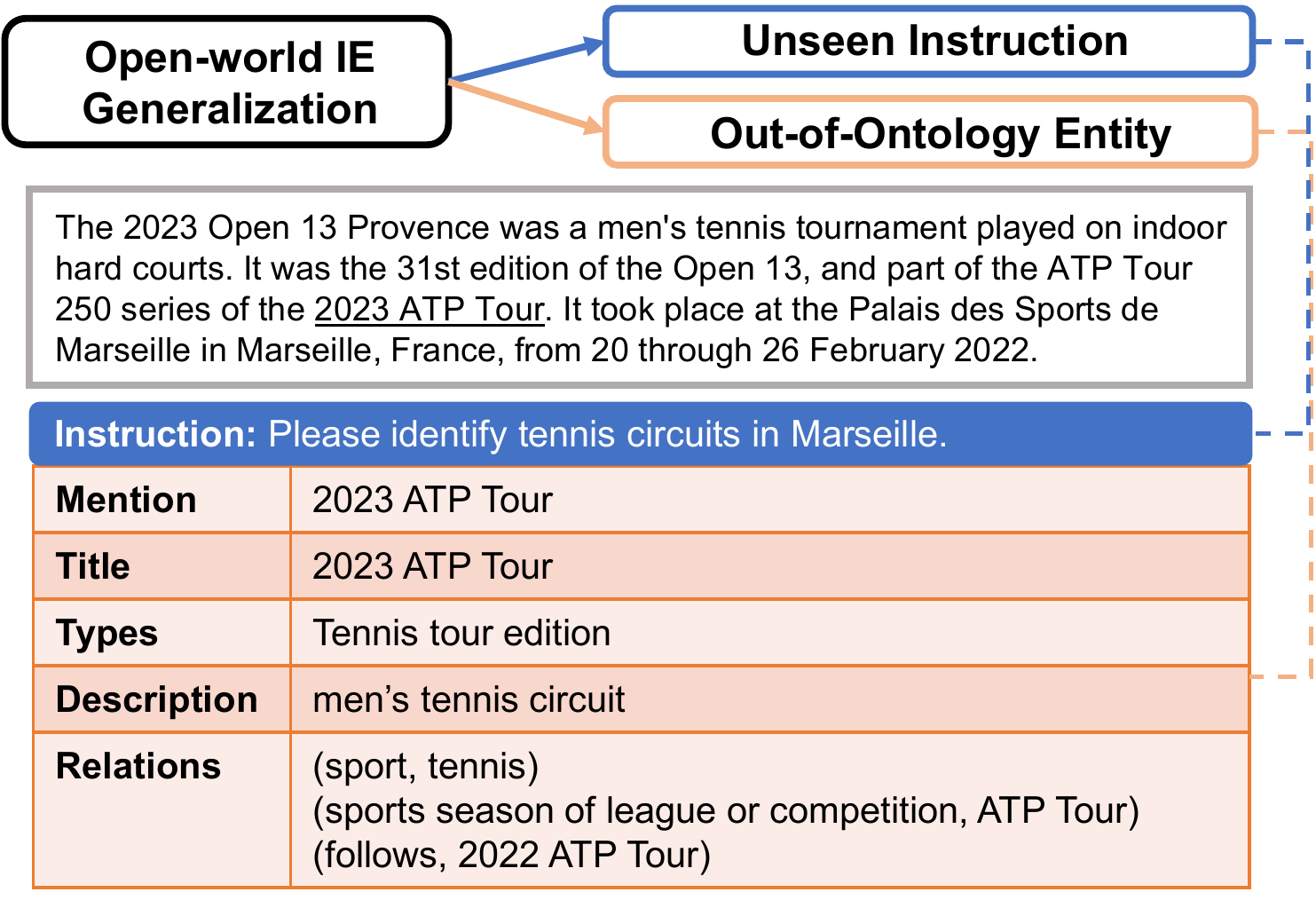}
    \caption{
        Illustration of Open-world IE and its two main challenges: generalization to unseen instructions and out-of-ontology entities.
    }
    \label{fig:teaser}
    \vspace{-1.5em}
\end{figure}

Most existing IE methods center around a \textbf{closed-world} setting with predefined ontologies.
For instance, NER generally extracts named entities within several categories~\cite{lample-etal-2016-neural};
EL focuses on associating mentions with a predefined ontology~\cite{kolitsas-etal-2018-end}.
Furthermore, conventional closed-world IE usually extracts all information without focusing on desired targets.
To better address these issues, we introduce Open-world Information Extraction (Open-world IE) to accommodate broad and diverse requests related to entity profiles surpassing predefined ontologies' limits.
Specifically, Open-world IE accepts an unstructured corpus and an instruction that characterizes target entities, identifies all entities within the context, and generates entity profiles, as shown in \Cref{fig:teaser}.
Open-world IE aims to break the ontology limitations in traditional IE and pursuit generalization to unseen instructions and out-of-ontology cases.
Past research on this topic has predominantly focused on individual subtasks of IE, such as EL~\cite{iurshina2022nilk,ruas2022nilinker} and OpenIE~\cite{niklaus-etal-2018-survey,bhardwaj-etal-2019-carb}.
Consequently, a noticeable gap exists in comprehensive end-to-end studies aiming to create more extensive entity profiles within an open-world setting.

With the emergence of large language models~(LLMs)~\cite{zhao2023survey}, generative IE based on LLMs holds substantial promise in addressing this open-world challenge, given their exceptional generalization capabilities.
Open-world IE can also serve as a pivotal capability for integrating plugins into the ChatGPT system, since it provides a flexible communication interface between LLMs and their plugins.
Nevertheless, existing research on LLMs reveals that they typically do not function as zero-shot learners in IE, necessitating appropriate instruction tuning to enhance their IE capabilities~\cite{ma2023large,wadhwa2023revisiting}.
Therefore, instruction tuning~\cite{wei2022finetuned} becomes critical in endowing LLMs with Open-world IE abilities.

To combat these issues, we develop \modelnamewex \modelnameex.
\modelname is an LLM designed for Open-world IE.
We formulate Open-world IE as an instruction-following auto-regressive generative task to generate comprehensive entity profiles in JSON.
We cover eight popular categories of instructions in various granularities.
Each category of instruction imposes specific constraints on candidate entities.
In pursuit of generalization over unseen instructions and out-of-ontology entities, we develop an instruction tuning dataset \datasetname for Open-world IE, which includes diverse instructions that endows \modelname with strong instruction following capability.
\datasetname incorporates rich entity and relation annotations, various instructions, and a delicate design of the out-of-ontology evaluation set, which contributes to the generalization of both unseen instructions and out-of-ontology entities.

The contributions of this work are three-fold.
First, we propose the definition of open-world IE and develop \modelname, which performs IE without the limitations of predefined ontology.
This flexibility allows for its generalization abilities and application across diverse downstream scenarios.
Second, we construct a substantial Open-world IE dataset \datasetname.
Third, we explore a comprehensive evaluation for Open-world IE.
We meticulously design an open-world evaluation set incorporated in \datasetname to assess Open-world IE capabilities thoroughly, focusing on the generalization of unseen instructions and out-of-ontology entities.
Our contributions are verified with experiments and multifaceted analysis.
Most notably, \modelname exhibits impressive generalization capabilities on unseen instructions and out-of-ontology cases, demonstrating its robust potential to address the open-world challenge effectively.

\section{Related Works}
\xhdr{Large Language Models}
Large language models~(LLMs) is an emerging topic summarized in a recent survey \citet{zhao2023survey}.
Therefore, we only provide a highly-selective review.
\citet{brown2020language} train an auto-regressive language model \textsc{GPT-3} with 175 billion parameters, showing extraordinary task-agnostic few-shot performance.
\citet{chowdhery2022palm} develop a Pathways Language Model \textsc{PaLM} and scale it up to 540 billion parameters.
\citet{scao2022bloom} propose \textsc{BLOOM}, open-access LLMs from 560 million to 175 billion parameters.
\citet{touvron2023llama} develop \textsc{LLaMa}, a more efficient public-accessible LLM.
We use \textsc{BLOOM} as the backbone since it was the latest public LLM pretrained on a diverse corpus, including codes.
However, other latest LLMs, such as \textsc{LLaMa}, can also be easily tuned on our dataset to acquire open-world IE abilities.

\xhdr{Instruction Tuning}
Instruction tuning is an emergent paradigm that finetunes LLMs on datasets described by instructions.
\citet{wei2022finetuned} finetune an LLM with 175 billion parameters on various NLP datasets with instruction templates and proof instruction tuning can significantly improve zero-shot performance.
\citet{ouyang2022training} show supervised instruction tuning and finetuning with human feedback helps LLMs align with human intent.
This work is further extended by \textsc{OpenAI} and becomes the product \textsc{ChatGPT}\footnote{\url{https://openai.com/blog/chatgpt}} used as a baseline in our work.
In this work, we create an instruction-following dataset \datasetname for open-world IE and employ instruction tuning to empower LLMs with Open-world IE abilities.

\xhdr{Information Extraction}
Instruction-following IE reformulates IE into a generation task with instructions describing target information.
We mainly present two concurrent works as this is an emerging topic.
\citet{wei2023zero} solve IE as a multi-turn question-answering format by providing predefined instructions to \textsc{ChatGPT}.
\citet{wang2023instructuie} proposes an instruction-tuning IE benchmark and develops a unified IE method.
However, all these works are based on the closed-world setting and have not adapted to Open-world IE, which is exactly our focus in this work.
To our best knowledge, \modelname is the first work exploring instruction-following open-world IE.
Previous explorations are limited to different sub-fields of IE, such as the NIL problem in EL~\cite{lin2012no} and open information extraction \cite{zhou2022survey}.
Open-world knowledge graph completion (KGC)~\cite{ye2022generative} also completes existing KGs by creating unseen entities and conducting link prediction to existing entities.
However, open-world KGC mainly uses structured information within KGs while we leverage rich unstructured corpus.

\begin{figure*}[t]
    \centering
    \includegraphics[width=\linewidth]{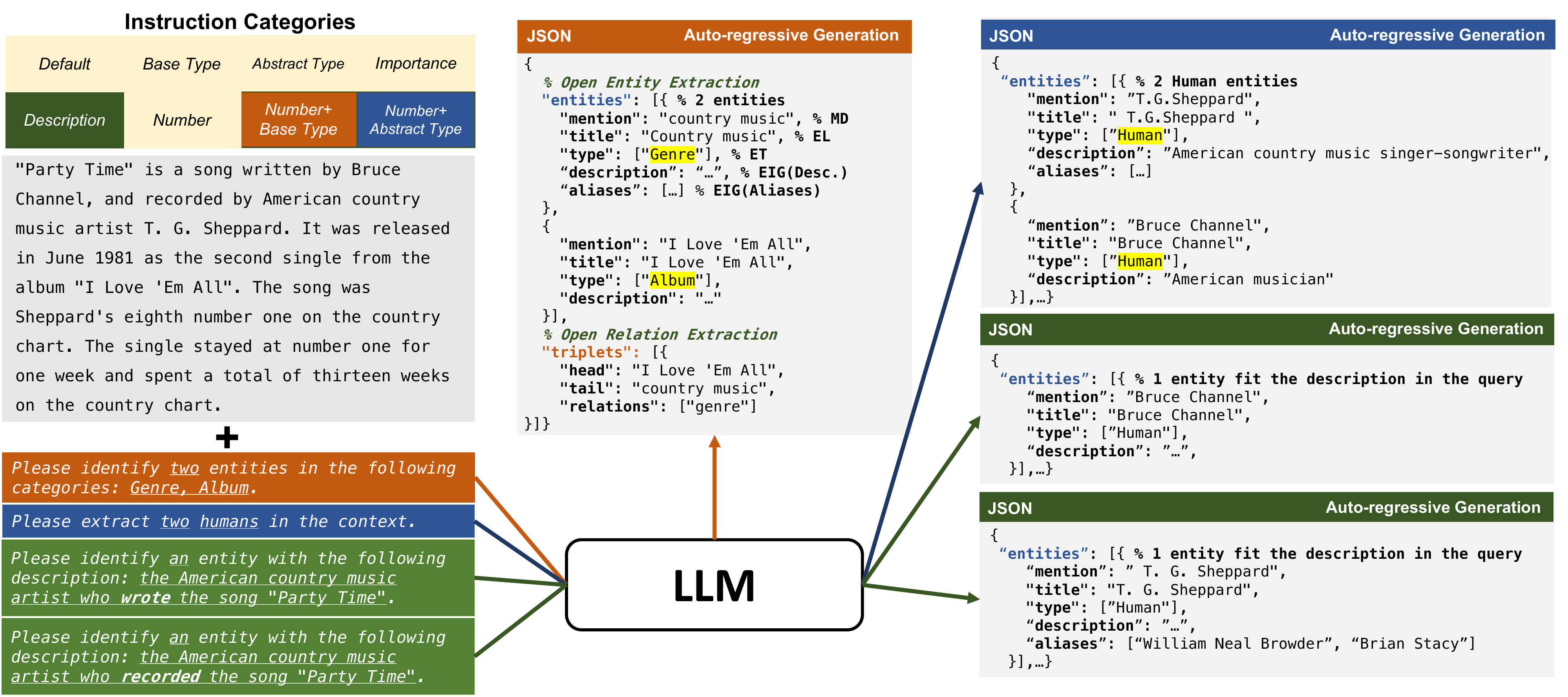}
    \caption{Overview of the open-world information extraction method \modelname.
    This figure shows four generation cases of \modelname under three instruction categories colored in orange, blue, and green.
    \modelname takes the corpus and instructions as inputs and auto-regressively generates JSON sequences.
    And \modelname can extract different entity profiles based on different instructions from the same corpus.
    The auto-regressive generation of the JSON targets aligns well with various IE subtasks marked as green comments in the JSON.
    }
    \label{fig:method}
    \vspace{-1em}
\end{figure*}

\begin{figure*}[t]
    \centering
    \includegraphics[width=\linewidth]{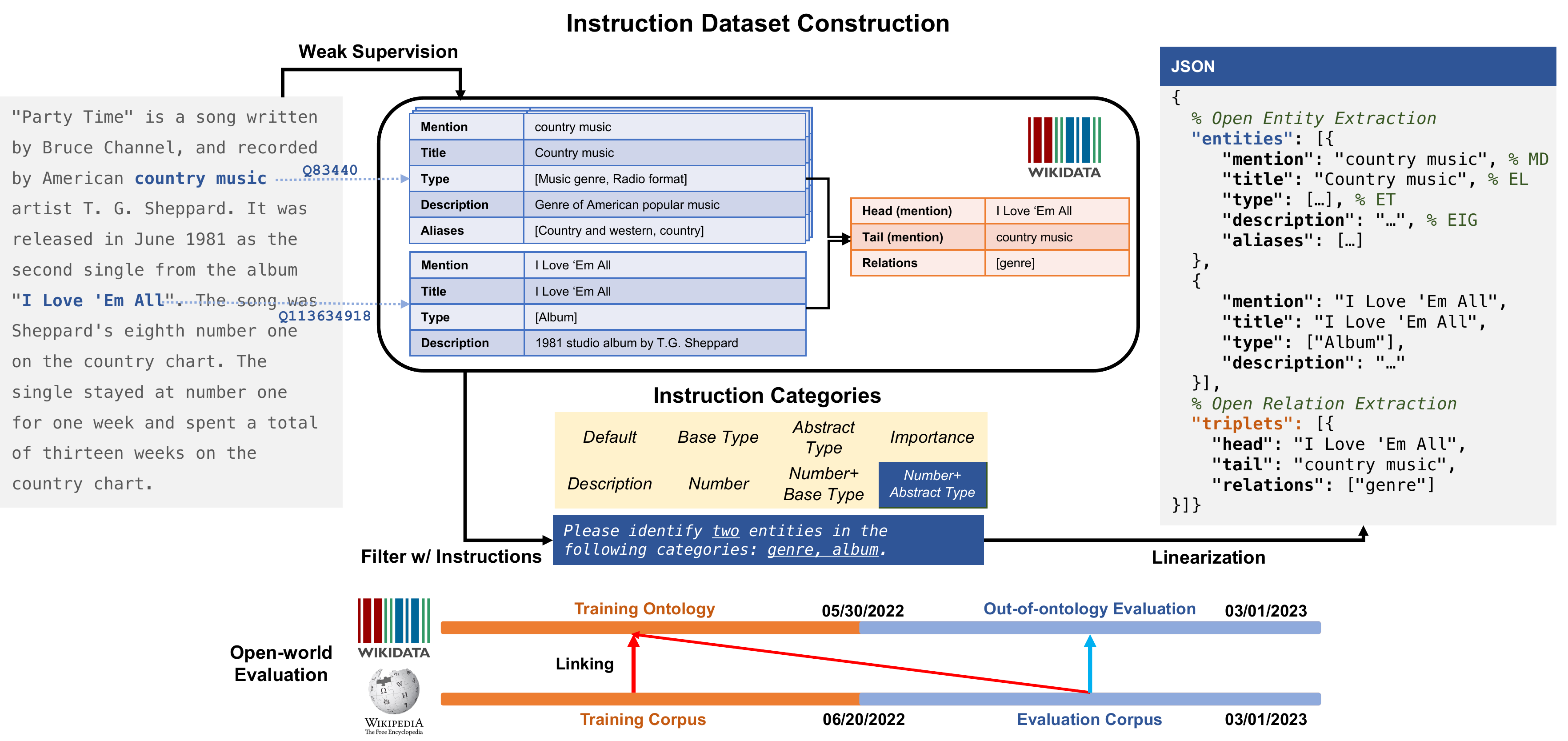}
    \caption{
    Overview of instruction-following open-world IE dataset construction.
    The top of this figure shows \datasetname is created by aligning anchor links in Wikipedia to entities and relations in the knowledge base Wikidata, and then augmented with instructions within eight categories.
    Entity profiles are identified from Wikidata and linearized into a sequence in the JSON format.
    The bottom of this figure shows how we create an open-world evaluation set with the time difference between Wiki dumps.
    }
    \label{fig:dataset-construction}
    \vspace{-1em}
\end{figure*}

\section{Methods}
In this section, we describe \modelname \modelnameex, an open-world generative IE method with instruction-following abilities.
We will introduce the preliminaries~(\Cref{sec:preliminaries}), instruction-following open-world IE~(\Cref{sec:instruction-following-open-world-IE}), and construction of the dataset \datasetname~(\Cref{sec:dataset-construction}).

\subsection{Preliminaries}\label{sec:preliminaries}

\xhdr{Problem Definition} Open-world IE aims to extract entity profiles from unstructured texts without predefined ontologies by following specific instructions.
In our task, the entity profile includes an entity mention, a canonical title, types, a description, aliases, and relations, as shown in \Cref{fig:teaser}.
Specifically, mentions are text spans in input corpus that are linked to entities; types are a list of phrases that an entity is an instance of; aliases are a list of synonyms; relations are a list of relation titles between extracted entities within the input. 
Given a document and an instruction describing a specific constraint about target entities, such as types or descriptions, Open-world IE methods are expected to generate entity profiles that fixed constraints in the instruction.

\xhdr{Method Overview} We solve Open-world IE by instruction tuning of LLMs.
As shown in \Cref{fig:method}, we first reformulate Open-world IE into auto-regressive generation by linearizing the structure knowledge into the JSON format (\Cref{sec:instruction-following-open-world-IE}).
We apply instruction tuning to empower \modelname extracting different entities following instructions within eight instruction categories in different granularities.
To do so, we build \datasetname, a large-scale instruction-following open-world IE dataset.
As presented in \Cref{fig:dataset-construction}, \datasetname is created by weak supervision between large-scale corpus and existing knowledge base (\Cref{sec:dataset-construction}).
Then, we augment the dataset with diverse instructions and rephrase them to enrich semantic diversity.
We also comprehensively evaluate \modelname on the open-world evaluation set (\Cref{sec:evaluation}).

\subsection{Instruction Tuning for Open-world IE}\label{sec:instruction-following-open-world-IE}

Instruction tuning is a multi-task learning paradigm that uses instructions to prompt models to generate proper outputs.
Leveraging the strong generalization abilities of LLMs to pursue generalization on unseen instructions and out-of-ontology cases, we reformulate Open-world IE as an instruction-following generation task.
To do so, we create diverse instructions and linearize structured IE outputs into JSON sequences.
And then we finetune LLMs in a supervised setup of instruction tuning (SFT), training LLMs to generate the targeted JSON sequence of entity profiles as the output.

\xhdr{Instruction Design}
The diversity of instruction categories are essential for the generalization of unseen instructions.
We first manually designed eight instruction categories with varying granularities to introduce extraction constraints that one might face in real-world applications:
\begin{itemize}[leftmargin=1em]
    \vspace{-0.5em}
    \setlength\itemsep{-0.5em}
    \item \textbf{Default}: Extract all entities and relations in the input without any additional requirements.
    \item \textbf{Base Type}: Extract entities of given base types. We define base types as fine-grained types in Wikidata, building from the ``P31 (instance of)'' properties of entities in Wikidata. 
    \item \textbf{Abstract Type}: Extract entities of given abstract types. We define abstract types as more coarse-grained types obtained by finding parent ``P279 (subclass of)'' properties of base types in Wikidata.
    This instruction category is only designed for extending type semantics during the training and helps LLMs learn more about the type hierarchy.
    Both instructions with base and abstract types share the same instruction prompts so we do not distinguish them in inference.
    \item \textbf{Description}: Extract entities that fit given descriptions.
    This instruction category provides ultra fine-grained instructions that require open-world IE methods directly understand diverse descriptions.
    As shown in the green cases in \Cref{fig:method}, descriptions can be phrases or sentences that describe the properties of entities.
    \item \textbf{Importance}: Extract the top-K most important entities.
    The importance of entities is defined as entity priorities in Wikidata.
    This instruction requires Open-world IE methods to rank entities with inherent priorities properly and only provide the top-K results.
    \item \textbf{Number}: Extract a specific number of entities in the input document.
    Unlike other categories, instructions with number requires our method to extract partial information from the input and the answer is not unique.
    Therefore, we separately analyze these categories in evaluation.
    \item \textbf{Number+Base Type}: Cross instructions between categories \textbf{Number} and \textbf{Base Type}.
    \item \textbf{Number+Abstract Type}: Cross instructions between categories \textbf{Number} and \textbf{Abstract Type}.
    \vspace{-2em}
\end{itemize}
We manually design an instruction template for each category.
Then we ask \textsc{ChatGPT} to rephrase the manually designed templates, enhancing the semantic diversity of these templates.
Details of original and rephrased templates are shown in \Cref{tab:instruction}.
\textsc{ChatGPT} prompts we use for rephrasing seed templates are shown in \Cref{tab:rephrase-prompts}.
We train \modelname to follow the top six single instructions.
And we add the last two cross instructions in evaluation to evaluate generalization on unseen instructions.

\xhdr{Linearization}\label{sec:linearization}
Previously, various techniques have been explored to linearize the structured information in the generative IE~\cite{ye2022generative} but either lack semantics or require additional training for special tokens~\cite{lou2023universal,wang2021zero,lu2022unified}.
To better leverage pretrained knowledge, we present a novel idea that uses the JSON (\underline{J}ava\underline{S}cript \underline{O}bject \underline{N}otation) format to linearize heterogeneous structured entity profiles.
It is primarily used to transmit data in web applications, so it frequently appears in codes.
Therefore, LLMs pretrained on codes are familiar with the JSON schema, avoiding additional training for any special tokens or manually-defined structure templates.
Furthermore, JSON uses a text-based syntax with key-value pairs, capturing additional semantics in natural language by keys and providing flexible structures.
This linearization aggregates multiple IE subtasks, revealing the chain of thoughts in IE employed in traditional pipeline methods.

\subsection{Instruction Dataset Construction}\label{sec:dataset-construction}
Learning from a large-scale instruction tuning dataset with a rich corpus and large training ontology contributes to the generalization of out-of-ontology cases.
However, building a large-scale Open-world IE dataset by manual annotations is infeasible since identifying entities in text and linking them with entity profiles require tremendous human effort.
Therefore, we develop a weakly supervised method that automatically creates the dataset \datasetname for instruction tuning.

\xhdr{Weak Supervision}
Our dataset is created by aligning anchor links in Wikipedia\footnote{\url{https://www.wikipedia.org/}} to entity profiles in its corresponding knowledge base Wikidata\footnote{\url{https://www.wikidata.org}} by the wiki identifiers, shown in the left part of \Cref{fig:dataset-construction}.
Wikipedia is a large corpus covering various domains, while Wikidata contains rich world knowledge.
Wikipedia and Wikidata are frequently revised by contributors worldwide, ensuring precision and being up-to-date.
All anchor links in Wikipedia are manually annotated, so linking between mentions and entities is reliable.
We only use the leading paragraph in each Wikipedia article since it contains the richest anchor links.
Besides, anchor links referring to the same entity may only be marked once the first time within an article, so using the rest of the paragraphs will face higher risks of missing mention annotations.
We retrieve four fields from Wikidata as its profile for each linked entity, including the canonical title, types, description, and aliases.
Canonical titles are English labels of entities;
types of entities are derived from ``instance of (P31)'' properties.
After identifying all entities in a paragraph, we employ distant supervision to identify relations between these entities from the knowledge base of Wikidata as described at the top of \Cref{fig:dataset-construction}.
Specifically, we link a relation triplet in the KB to this paragraph if both head and tail entities are mentioned.
A relation triplet is represented by mentions of head and tail entities and a list of relation names.
The detailed statistics of this dataset are presented in \Cref{app:data-statistics}.

\xhdr{Instruction Augmentation}
We further augment the dataset with predefined instructions as shown in the middle of \Cref{fig:dataset-construction}.
We generate an instruction-tuning sample with the default instruction and randomly select one another from six training categories for each sample.
All instructions focus on entities, and we also filter out triplets whose head or tail mentions are filtered out during the augmentation to ensure alignment.
Specifically, we augment samples with abstract type instructions using parent ``P279'' to replace base types.

\section{Evaluation}\label{sec:evaluation}

Open-world IE focuses on extracting unseen out-of-ontology entities and relations.
Therefore, we create an open-world evaluation set with rich out-of-ontology cases and design metrics for evaluating such performance.

\xhdr{Open-world Evaluation Set}
A well-designed open-world evaluation set is essential for evaluating Open-world IE methods without bias.
Previous work constructs open-world test sets by simply holding out a portion of entities from the training ontology.
However, such methods may introduce potential risks and lead to insufficient evaluation.
First, holding out entities from the training ontology also removes corresponding links on mentions in the training corpus, hindering the completeness of annotations.
Moreover, mentions of these held-out entities still frequently appear in the training corpus even if they are not specifically annotated, resulting in potential data leakage.
In real-world cases, most emerging out-of-ontology entities are unpopular in the training corpus.
This held-out method can not achieve the original goal of evaluating generalization on such entities.
Therefore, we propose a delicate method that uses the time difference between Wiki dumps to construct a strictly open-world test set.
As shown in the bottom of \Cref{fig:dataset-construction}, we use the Wikidata dump on 05/30/2022 and the Wikipedia dump on 06/20/2022 to build the training set.
As the evaluation corpus, we filter all new articles between two Wikipedia dumps, 06/20/2022 and 03/01/2023.
We also select new entities appearing in the Wikidata dump on 03/01/2023 as out-of-ontology entities, which are not presented in the Wikidata on 05/30/2022.
The \textsc{ROOTS} corpus~\cite{laurenon2022the}, pretrained corpus of \textsc{BLOOM}, only includes Wikipedia dump before 06/20/2022, so this evaluation set also remains unseen for the \textsc{BLOOM} pretraining.
Using time difference to build an out-of-ontology evaluation set minimizes potential data leakage and maximizes the completeness of the training set's annotations.

\xhdr{Metrics}
Although defining Open-world IE as an end-to-end entity profile generation task, we still split it into six tasks in evaluation to provide more comprehensive analyses:
\textbf{(1) Mention Detection~(MD)} corresponds to the correctness of the ``mention'' key in the JSON output.
\textbf{(2) Entity Linking~(EL)} is related to the ``title'' key, evaluating whether models generate proper canonical titles for mentions.
We use hard and soft matching based on a ROUGE-L F1 threshold as the criterion.
\textbf{(3) Entity Typing~(ET)} requires models generate entity types for entities.
\textbf{(4) Open Relation Extraction~(RE)} is related to the ``triplets'' field in the JSON output.
We learn from OpenIE evaluation~\cite{zhou2022survey} and use CaRB~\cite{bhardwaj2019carb} to evaluate triplet generation performance.
We calculate metrics based on CaRB with the ROUGE-L matcher.
\textbf{(5) Description Generation~(EIG-Desc.)} requires models to generate description for generated entities. We report the average ROUGE-L F1 for this task.
\textbf{(6) Aliases Generation~(EIG-Aliases)} is related to the ``aliases'' field, expecting models to generate aliases for predicted entities.
We report precision, recall, and F1 scores on each task except description generation.
We randomly select three rephrased templates for each sample in the test set and report the average metric with standard deviation.

\xhdr{Unseen Ontologies}
We explore the out-of-ontology generalization by separately analyzing the recall of training~(Before 05/30/2022) and out-of-training~(After 05/30/2022) entities.
For instance, \textit{2023 ATP Tour~(Q111441127)} shown in \Cref{fig:teaser} is an unseen entity introduced to Wikidata after 05/20/2022.
Open-world IE methods are proven to have the great generalization of unseen ontologies if they can extract this entity from the latest corpus.

\xhdr{Unseen Instructions}
We also split the test set into samples with unseen and seen instructions under the most fine-grained category ``Description''.
Unseen instructions query context with constraints that are not in the training instructions.
For example, ``men's tennis circuit'' is a description not shown in any training instructions.
So the instruction extracting entities with such description is considered as an unseen instruction.
Similarly, we filter out unseen instructions for other categories and separately evaluate Open-world IE performance on this split.
The unseen proportions in each category are shown in \Cref{tab:data-statistics-instruction-split}. 
We also separately evaluate number-related instructions as these partial extraction instructions have no unique correct answers.

\section{Experiments}
\begin{table*}[t]
    \centering
    \small
    \setlength{\tabcolsep}{2.4mm}{
    \begin{tabular}{cccccccc}
    \toprule
    \multirow{2}{*}{\textbf{Method}} & \textbf{MD} & \multicolumn{2}{c}{\textbf{EL}} & \textbf{ET} & \textbf{OpenRE} & \textbf{EIG~(Desc.)} & \textbf{EIG~(Aliases)} \\
    & F1 & F1(T=1) & F1(T=0.8) & F1 & F1(CaRB) & F1(ROUGE-L) & F1 \\
    \midrule
    \textsc{GENRE} & $43.7^{\dag}$ & $17.2^{\dag}$ & $20.1^{\dag}$ & $-$ & $-$ & $-$ & $-$ \\
    \textsc{OpenIE6} & $-$ & $-$ & $-$ & $-$ & $15.2^{\dag}$ & $-$ & $-$ \\
    \midrule
    \textsc{ChatGPT} & $50.5_{0.2}$ & $25.0_{0.2}$ & $26.1_{0.2}$ & $7.9_{0.0}$ & $22.5_{0.2}$ & $39.0_{0.2}$ & $15.8_{0.0}$ \\
    \textsc{ChatGPT w/Demo} & $51.1_{0.1}$ & $38.6_{0.0}$ & $39.9_{0.0}$ & $14.8_{0.1}$ & $21.4_{0.0}$ & $52.0_{0.2}$ & $13.0_{0.1}$ \\
    \modelname-1b & $61.4_{0.0}$ & $49.6_{0.0}$ & $50.6_{0.0}$ & $40.4_{0.0}$ & $56.1_{0.2}$ & $68.6_{0.0}$ & $72.7_{0.1}$ \\
    \modelname-7b & $\mathbf{79.6_{0.0}}$ & $\mathbf{69.8_{0.1}}$ & $\mathbf{70.7_{0.1}}$ & $\mathbf{56.4_{0.0}}$ & $\mathbf{67.8_{0.2}}$ & $\mathbf{80.2_{0.1}}$ & $\mathbf{80.5_{0.0}}$ \\
    \bottomrule
    \end{tabular}}
    \caption{Main results of overall performance in end-to-end evaluation.
    We report the macro average of F1 scores with all instruction categories on mention detection~(MD), entity linking~(EL), entity typing~(ET), open relation extraction~(OpenRE), and entity information generation (EIG) for descriptions and aliases.
    \DevEx
    Comprehensive results are demonstrated in \Cref{app:comprehensive-results}.
     \textsuperscript{\dag} We only report the performance of \textsc{GENRE} and \textsc{OpenIE6} on default instructions since they are closed-world benchmarks focusing on specific subtasks without instruction following abilities.
    }
    \label{tab:main-results-overall-performance}
\end{table*}

\begin{table*}[t]
    \centering
    \small
    \setlength{\tabcolsep}{1.5mm}{
    \begin{tabular}{ccccccccc}
    \toprule
    \multirow{2}{*}{\textbf{Partition}} & \multirow{2}{*}{\textbf{Method}} & \textbf{MD} & \multicolumn{2}{c}{\textbf{EL}} & \textbf{ET} & \textbf{OpenRE} & \textbf{EIG~(Desc.)} & \textbf{EIG~(Aliases)} \\
    & & R & R(T=1) & R(T=0.8) & R & R(CaRB) & R(ROUGE-L) & R \\
    \midrule
    \multirow{6}{*}{\makecell[c]{Before\\05/30/2022}} & \textsc{GENRE} & $55.3^{\dag}$ & $23.8^{\dag}$ & $24.7^{\dag}$ & $-$ & $-$ & $-$ & $-$ \\
    & \textsc{OpenIE6} & $-$ & $-$ & $-$ & $-$ & $21.9^{\dag}$ & $-$ & $-$ \\
    \cline{2-9}
     & \textsc{ChatGPT} & $59.3_{0.3}$ & $31.8_{0.2}$ & $32.3_{0.2}$ & $9.7_{0.1}$ & $34.1_{0.2}$ & $40.2_{0.4}$ & $10.1_{0.1}$ \\
	 & \textsc{ChatGPT w/Demo} & $56.5_{0.0}$ & $45.0_{0.1}$ & $45.7_{0.1}$ & $15.9_{0.2}$ & $38.3_{0.0}$ & $53.6_{0.2}$ & $8.4_{0.0}$ \\
	 & \modelname-1b & $59.2_{0.1}$ & $51.8_{0.0}$ & $52.2_{0.0}$ & $44.4_{0.1}$ & $61.8_{0.3}$ & $71.4_{0.1}$ & $69.7_{0.1}$ \\
	 & \modelname-7b & $\mathbf{83.7_{0.0}}$ & $\mathbf{78.8_{0.1}}$ & $\mathbf{79.1_{0.1}}$ & $\mathbf{66.8_{0.1}}$ & $\mathbf{79.4_{0.1}}$ & $\mathbf{82.5_{0.0}}$ & $\mathbf{78.0_{0.0}}$ \\
    \midrule
    \multirow{6}{*}{\makecell[c]{After\\05/30/2022}} & \textsc{GENRE} & $33.6^{\dag}$ & $0^{\dag}$ & $3.6^{\dag}$ & $-$ & $-$ & $-$ & $-$ \\
    & \textsc{OpenIE6} & $-$ & $-$ & $-$ & $-$ & $18.2^{\dag}$ & $-$ & $-$ \\
    \cline{2-9}
     & \textsc{ChatGPT} & $58.9_{0.2}$ & $23.8_{0.2}$ & $26.5_{0.3}$ & $6.9_{0.1}$ & $27.1_{0.5}$ & $36.0_{0.3}$ & $14.3_{0.3}$ \\
	 & \textsc{ChatGPT w/Demo} & $58.2_{0.0}$ & $39.1_{0.0}$ & $42.1_{0.0}$ & $14.2_{0.1}$ & $29.3_{0.5}$ & $48.1_{0.2}$ & $13.3_{0.1}$ \\
	 & \modelname-1b & $55.1_{0.0}$ & $36.3_{0.1}$ & $38.3_{0.1}$ & $22.8_{0.0}$ & $34.5_{0.1}$ & $55.7_{0.3}$ & $20.5_{0.2}$ \\
	 & \modelname-7b & $\mathbf{71.7_{0.1}}$ & $\mathbf{51.7_{0.2}}$ & $\mathbf{53.8_{0.2}}$ & $\mathbf{24.1_{0.2}}$ & $\mathbf{36.1_{0.1}}$ & $\mathbf{64.1_{0.2}}$ & $\mathbf{22.7_{0.1}}$ \\
    \bottomrule
    \end{tabular}}
    \caption{Main results of generalization analysis in end-to-end evaluation.
    Headers are the same as \Cref{tab:main-results-overall-performance} except we report recalls in each task.
    \PartitionEx
    Comprehensive results are demonstrated in \Cref{app:comprehensive-results}.
    \textsuperscript{\dag} We only report the performance of \textsc{GENRE} and \textsc{OpenIE6} on default instructions since they are closed-world benchmarks focusing on specific subtasks without instruction following abilities.
    }
    \label{tab:main-results-generlization-study}
    \vspace{-1em}
\end{table*}

\begin{figure*}[t]
    \centering
    \includegraphics[width=\linewidth]{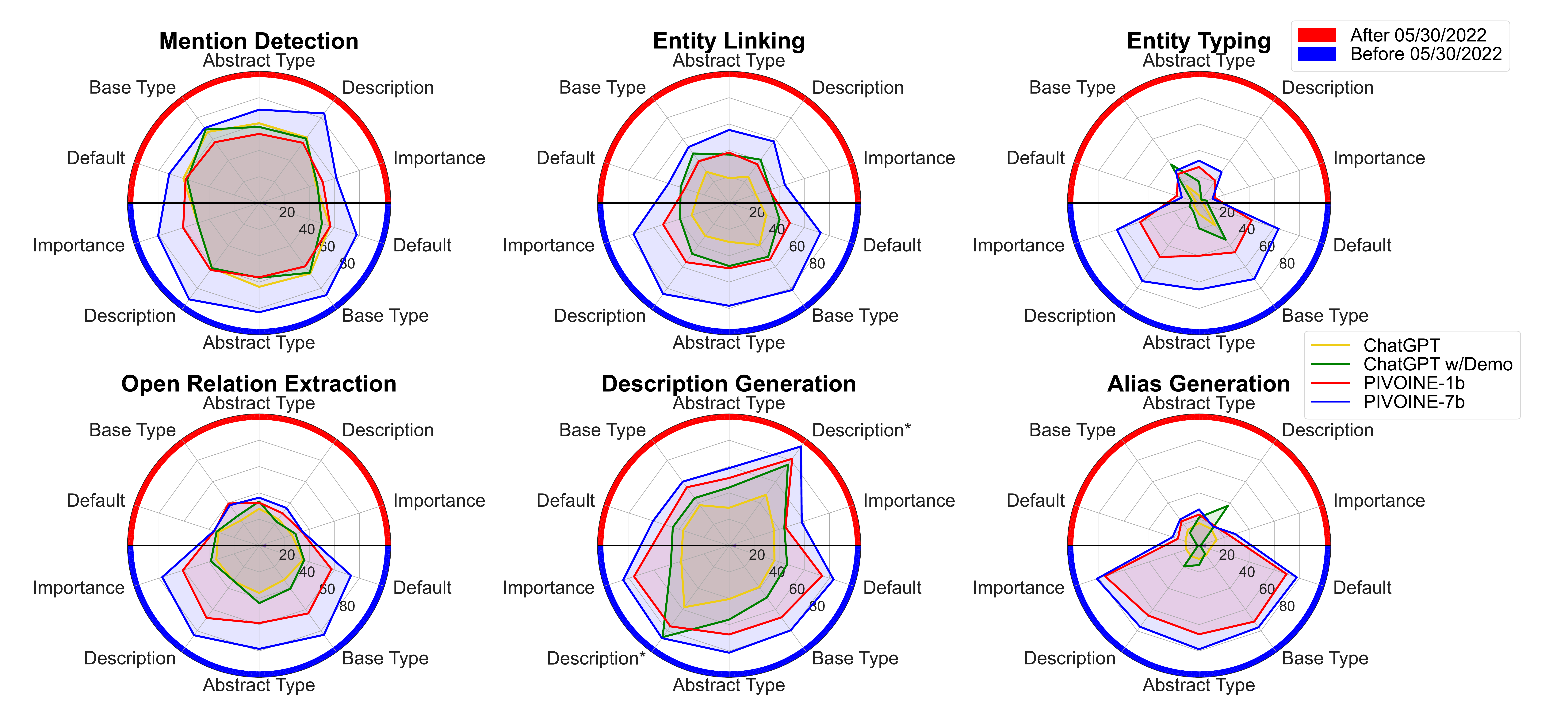}
    \caption{
    Main results of the \textbf{ontology generalization} evaluation for each instruction category in each task.
    \PartitionEx
    \textsuperscript{*} To avoid data leakage, descriptions as instructions are not considered in description generation tasks.
    Original scores are shown in \Cref{app:comprehensive-results}.
    }
    \label{fig:main-results-instruct}
\end{figure*}

\begin{table*}[t]
    \centering
    \small
    \setlength{\tabcolsep}{1.5mm}{
    \begin{tabular}{c|c|ccccccc}
    \toprule
    \multirow{2}{*}{\textbf{Method}} & \multirow{2}{*}{\textbf{Partition}} &\textbf{MD} & \multicolumn{2}{c}{\textbf{EL}} & \textbf{ET} & \textbf{OpenRE} & \textbf{EIG~(Aliases)} \\
    & & F1 & F1(T=1) & F1(T=0.8) & F1 & F1(CaRB) & F1 \\
    \midrule
	\multirow{2}{*}{\textsc{ChatGPT}} & unseen  & $60.3_{0.6}$ & $24.7_{1.1}$ & $26.6_{1.2}$ & $3.0_{0.4}$ & $24.9_{0.4}$ & $12.1_{1.4}$ \\
	 & seen  & $67.4_{0.9}$ & $25.1_{0.3}$ & $28.8_{0.8}$ & $1.0_{0.7}$ & $23.1_{1.4}$ & $22.9_{3.2}$ \\
    \midrule
	\multirow{2}{*}{\textsc{ChatGPT w/Demo}} & unseen  & $59.0_{0.2}$ & $40.0_{0.5}$ & $41.9_{0.6}$ & $2.3_{0.3}$ & $22.2_{0.6}$ & $22.6_{1.8}$ \\
	 & seen  & $67.9_{1.8}$ & $44.3_{2.6}$ & $48.2_{2.8}$ & $7.1_{0.0}$ & $23.3_{0.7}$ & $\mathbf{74.2_{3.8}}$ \\
    \midrule
	\multirow{2}{*}{\modelname-1b} & unseen  & $55.8_{0.4}$ & $34.7_{0.3}$ & $36.4_{0.3}$ & $19.9_{0.1}$ & $30.0_{0.6}$ & $9.7_{0.3}$ \\
	 & seen  & $60.3_{0.7}$ & $44.9_{0.7}$ & $47.5_{0.8}$ & $26.7_{0.7}$ & $33.9_{0.5}$ & $36.5_{0.4}$ \\
    \midrule
	\multirow{2}{*}{\modelname-7b} & unseen  & $\mathbf{83.6_{0.2}}$ & $\mathbf{55.1_{0.1}}$ & $\mathbf{57.8_{0.1}}$ & $\mathbf{28.2_{0.1}}$ & $\mathbf{35.0_{0.3}}$ & $\mathbf{12.8_{0.3}}$ \\
	 & seen  & $\mathbf{86.5_{0.2}}$ & $\mathbf{71.7_{0.2}}$ & $\mathbf{72.8_{0.2}}$ & $\mathbf{33.8_{0.7}}$ & $\mathbf{39.4_{0.0}}$ & $31.8_{0.3}$ \\
    \bottomrule
    \end{tabular}}
    \caption{Main results of the instruction generalization evaluation on ``Description'' instruction category.
    Headers are the same as \Cref{tab:main-results-overall-performance}.
    Partition denotes the unseen and seen descriptions in instructions.
    To avoid data leakage, descriptions as instrcutions are not considered in descirption generation tasks.
    \DevEx
    The best scores in each partition are marked in \textbf{bold}.
    }
    \label{tab:main-results-instruction-generalization-main}
\end{table*}

\begin{figure*}[t]
    \centering
  \includegraphics[width=\linewidth]{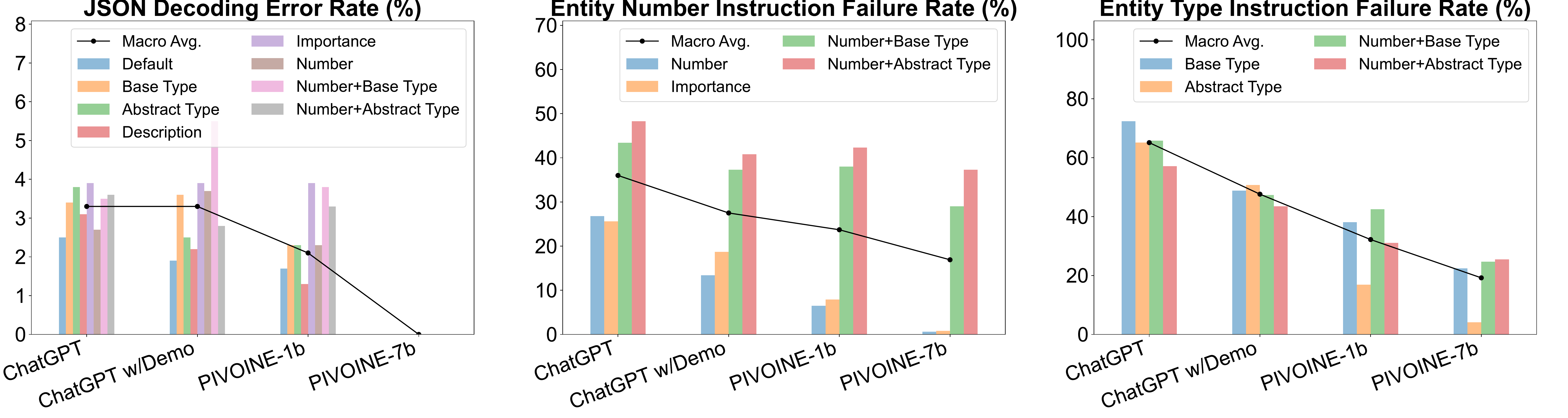}
    \caption{
    Analysis of the instruction-following capabilities, including JSON decoding error rate, entity number instruction failure rate, and entity type instruction failure rate.
    Original scores are shown in \Cref{app:comprehensive-results}.
    }
    \label{fig:analysis-instruction-following}
    \vspace{-1em}
\end{figure*}
In this section, we carry out comprehensive evaluation of our proposed \modelname. Specifically,  we present experimental setup~(\Cref{sec:experiment-setup}), main results~(\Cref{sec:results}), and further analysis of \modelname~(\Cref{sec:analysis}).

\subsection{Experimental Setup}\label{sec:experiment-setup}

\xhdr{Baselines} 
We employ ChatGPT as our main baseline since no instruction-following open-world IE methods exist to our best knowledge.
\textbf{(1) \textsc{ChatGPT}} is an instruction-following LLM that can handle various tasks.
The detailed configuration of \textsc{ChatGPT} is described in \Cref{app:chatgpt-configuration}.
\textbf{(2) \textsc{ChatGPT w/Demo}} is a stronger baseline with a one-shot demo based on the \textsc{ChatGPT} baseline.
We provide \textsc{ChatGPT} an example of the instruction from the same category by adding it to the conversation history.
We also introduce two traditional IE baselines to provide more comparisons of IE performance:
\textbf{(3) \textsc{GENRE}}~\cite{kumar-bojar-2022-genre} is the first system retrieving entities by generating canonical names, which can address mention detection and entity linking but is constrained by the KILT entities.
\textbf{(4) \textsc{OpenIE6}}~\cite{kolluru-etal-2020-openie6} is a recent state-of-the-art neural open information extraction system.
We employ it as a baseline of open relation extraction.

\xhdr{Configurations} We start from \textsc{BLOOM}~\cite{scao2022bloom} checkpoints with 1 billion and 7 billion parameters and run instruction tuning on \datasetname.
We use suggested hyper-parameters for finetuning each model in \citet{scao2022bloom}.
\modelname-1b is trained on 64 NVIDIA V100 GPU for 92 hours.
\modelname-7b is trained on 256 NVIDIA V100 GPU for 54 hours.
We develop our models with Megatron-Deepspeed from BigScience.
We trained \modelname-1b for 10,294 steps with a global batch size of 1,024 and \modelname-7b for 5,000 steps with a global batch size of 2,048.
The training steps are selected by evaluation of performance on the development set.
We infer on 256 NVIDIA V100 GPU within 30 minutes with generation parameters in \Cref{app:hyper-parameters}.

\subsection{Main Results}\label{sec:results}
We present our main results in three aspects: overall performance, generalization study on entities within or out of the training ontology, and unseen instructions.

\xhdr{Overall Performance}
\Cref{tab:main-results-overall-performance} shows overall performance on six subtasks.
We report the macro average of F1 scores with all instruction categories.
We only report performance on default instruction for \textsc{GENRE} and \textsc{OpenIE6} for reference since they have no instruction-following abilities.
They fail to achieve outstanding performance on corresponding tasks, showing Open-world IE is challenging for traditional IE methods.
\textsc{ChatGPT} can partially address Open-world IE but only has 7.9\% F1 score in ET and 15.8\% F1 score in EIG(Aliases).
\textsc{ChatGPT w/Demo} significantly outperforms \textsc{ChatGPT} in EL and description generation and have comparable performance on other tasks, showing the demo in history benefits \textsc{ChatGPT} on open-world IE.
\modelname-1b further outperforms \textsc{ChatGPT w/Demo} by about 10\% absolute improvement in F1 over all tasks.
\modelname-7b achieves the best performance among all methods, significantly outperforming \modelname-1b by nearly 20\% F1 in most tasks.
This result suggests larger models will boost the performance and even can potentially address some of the tasks.

\xhdr{Generalization to unseen ontologies}
To further evaluate generalization abilities on out-of-ontology cases, we separately analyze recalls of entities before and after 05/30/2022 in the open-world test set and present results in \Cref{tab:main-results-generlization-study}.
The partition ``Before 05/30/2022'' denotes mentions linked to the training ontology (seen entities) while ``After 05/30/2022'' denotes those that are out of training ontology (unseen entities).
We first witness a consistent performance drop on unseen entities for all methods, especially the closed-world baselines.
Even for the two ChatGPT-based baselines, performance on EL, ET, OpenRE, and EIG(Desc.) also drop dramatically, probably because unseen entities are the latest and unpopular.
\modelname-1b only outperforms \textsc{ChatGPT w/Demo} on four tasks in the unseen partition.
At the same time, it surpasses \textsc{ChatGPT w/Demo} on all tasks in the seen partition, suggesting such generalization requires a sufficient model scale.
\modelname-7b still achieves the best performance on tasks in both partitions, showing it can successfully generalize to out-of-ontology entities in Open-world IE.

\Cref{fig:main-results-instruct} shows a more detailed analysis of each instruction category in each task.
Comparing six radar charts, we identify MD, EL, and EIG(Desc.) are three tasks on which all methods are easy to generalize to out-of-ontology entities since their upper and lower parts of plots are nearly symmetric.
We notice such generalization is harder in RE since it is an end-to-end task requiring precise out-of-ontology MD first.
ET and EIG(Aliases) are the most challenging for all methods because they require a comprehensive list of types and aliases.
As for instruction categories, we find \modelname-7b consistently outperforms other baselines on all instruction categories, except for the Base Type instructions in ET and the Description instructions in EIG(Aliases).

\xhdr{Generalization to unseen instructions}
We evaluate the generalization abilities of both unseen instructions and out-of-ontology cases and present the results in \Cref{tab:main-results-instruction-generalization-main}.
As introduced in the last paragraph of \Cref{sec:evaluation}, unseen instructions include descriptions that have not appeared in the training samples.
We notice \modelname-7b achieves the best performance on almost all tasks under both seen and unseen partitions except for EIG(Aliases) in the unseen partition, where \textsc{ChatGPT w/Demo} outperforms other methods.
Therefore, \modelname-7b shows extraordinary generalization abilities to two main challenges in Open-world IE, especially in MD and EL.

\subsection{Analysis}\label{sec:analysis}

We provide further analyses to reveal model behaviors on instruction-following open-world IE.

\xhdr{Instruction Following}
We analyze instruction following qualities in three aspects as shown in \Cref{fig:analysis-instruction-following}.
First, we analyze the JSON decoding error rates with each method on different instructions.
Generating correct JSON schema is the essential requirement for parsing extracted information.
As presented in the left figure in \Cref{fig:analysis-instruction-following}, ChatGPT-based methods fail to ensure the valid JSON schema, especially on unseen cross instructions.
\modelname-1b is better than baselines on average but still faces decoding errors.
However, \modelname-7b has no JSON decoding errors on all instruction categories, even for unseen cross instructions.
Therefore, \modelname-7b is a trustworthy model consistently generating valid JSON schema.

We also compare entity number instruction failure rates on four number-related instruction categories.
The results are shown in the middle of \Cref{fig:analysis-instruction-following}.
\modelname-7b still has the lowest failure rates for extracting the correct number of entities, which is close to zero on two trained instruction categories.
All methods fail to follow number instructions when receiving untrained cross instructions, suggesting cross instructions are still challenging for current methods.
Furthermore, results in \Cref{tab:main-results-partial-md-ie} shows that number-related instructions do not hinder the precision of MD and EL.
These instruction categories provide partial extraction abilities to \modelname so that we can control the precision-recall trade-off by specifying the number of entities we need in the instruction.

Similar patterns are observed in the entity type instruction failure rate analysis shown in the right of \Cref{fig:analysis-instruction-following}.
\modelname-7b only has half failure rates on cross instruction categories compared with the vanilla \textsc{ChatGPT}, showing \modelname-7b has much better type instruction following abilities.
We also notice that following instructions with abstract types is significantly easier for \modelname than fine-grained base types.

\xhdr{Human Evaluation}
\modelname may extract correct relation triplets that are out of the scope of existing Wikidata.
Therefore, we randomly select 100 relation triplets with unseen entities predicted by \modelname-7b to analyze its precision further.
We consider the evidence of a relation triplet as the context from which \modelname-7b extracts the triplet.
To minimize the manual effort, we first reformulate the output of \modelname-7b to a prompt and ask GPT-4 to provide judgment and explanation based on evidence supplied by \modelname.
We also manually evaluate the correctness of relation triplets based on the same evidence without additional world knowledge.
The accuracy of relation triplets provided by GPT-4 is 87\%, and manual checking accuracy is 83\%.
The agreement between GPT-4 and the annotator is 87\%, suggesting GPT-4 is capable of evaluating relation correctness.
The latest Wikidata also verifies 8\% cases.
This evaluation shows \modelname-7b can precisely excavate new entities and relations from the corpus.

\section{Conclusion}
We propose Open-world IE, a challenging task that aims to extract out-of-ontology entity profiles with instructions.
Towards this grand mission, we create a large-scale instruction-following open-world IE dataset \datasetname and develop \modelname by instruction tuning.
We conduct extensive experiments on diverse instruction categories and different model scales, showing \modelname is a trustworthy LLM capable of following (possibly unseen) instructions in various granularities and extracting out-of-ontology entity profiles.
Valuable future works include extending \modelname to a larger scale and exploring a more comprehensive range of instruction categories.

\bibliography{anthology,custom}
\bibliographystyle{acl_natbib}

\appendix
\clearpage
\newpage

\section*{Appendix}

\section{Detailed Instructions}\label{app:detailed-instructions}

\Cref{tab:instruction} shows original manually-designed seed instruction, rephrased instructions, and number of rephrased instructions for each category.
We use ChatGPT to rephrased our original instructions.
The prompt we used is shown in \Cref{tab:rephrase-prompts}.

\section{Dataset Statistics}\label{app:data-statistics}

We display the statistics of \datasetname in \Cref{tab:data-statistics}.
\datasetname contains a rich corpus, which includes all head paragraphs of all Wikipedia articles with 39 million mentions and 19 million triplets.
And \datasetname is also annotated with a large ontology containing over 2 million entities with 21 thousand entity types and 962 relation types, ensuring it covers a wide range of domains.
Besides, the entity information density of \datasetname is also abundant, so models can be trained for extracting entity profiles efficiently.

\begin{table}[ht]
    \small
    \centering
    \begin{tabular}{cccc}
    \toprule
    Partition & Base Type & Abstract Type & Description \\
    \midrule
    Unseen & 18.9\% & 17.4\% & 82.3\% \\
    Seen & 81.1\% & 82.6\% & 17.7\% \\
    \bottomrule
    \end{tabular}
    \caption{Proportions of unseen and seen instructions in ``Base Type'', ``Abstract Type'', and ``Description'' in the test set.}
    \label{tab:data-statistics-instruction-split}
\end{table}

We also create a small development set to select the best checkpoint balancing performance on seen and unseen cases.
As shown in \Cref{tab:data-statistics}, the open-world evaluation test set contains rich unseen mentions~(27.1\%, 29,612/109,411) annotated by unseen entities~(24.2\%, 9,464/39,086).

\section{\textsc{ChatGPT} Configurations}\label{app:chatgpt-configuration}

We provide ChatGPT with the input context, instruction, and output JSON schema to prompt ChatGPT to solve instruction-following open-world IE.
We use ChatGPT in May 2022 and query it with the official API provided by OpenAI\footnote{\url{https://openai.com/blog/introducing-chatgpt-and-whisper-apis}}.
The detailed prompt we used is shown in \Cref{tab:chatgpt-open-world-ie-prompts}.

\section{Details of Human Evaluation}

We use the May 1st, 2023 version of GPT-4.
We query it via the chat platform of OpenAI\footnote{\url{https://chat.openai.com/}}.
The detailed cases are shown in our Github repository.

\section{Hyper-parameters}\label{app:hyper-parameters}
We infer \modelname with the official \textsc{BLOOM} inference project provided by HuggingFace\footnote{\url{https://github.com/huggingface/transformers-bloom-inference}}.
We generate with beam search without sampling and the number of beam is 4.
The maximum number of generated tokens are 2,048.
The other generation parameters are set to default.

\section{Comprehensive Results}\label{app:comprehensive-results}

\begin{table}[ht]
    \centering
    \small
    \setlength{\tabcolsep}{0.4mm}{
    \begin{tabular}{l|cccc}
    \toprule
    Instruction & ChatGPT & \makecell[c]{ChatGPT\\w/Demo} & \modelname-1b & \modelname-7b \\
    \midrule
    Default & $2.5_{0.1}$ & $1.9_{0.1}$ & $1.7_{0.0}$ & $0.0_{0.0}$ \\
	Base Type & $3.4_{0.3}$ & $3.6_{0.4}$ & $2.3_{0.2}$ & $0.0_{0.0}$ \\
	Abstract Type & $3.8_{0.4}$ & $2.5_{0.8}$ & $2.3_{0.2}$ & $0.0_{0.0}$ \\
	Description & $3.1_{0.3}$ & $2.2_{0.2}$ & $1.3_{0.3}$ & $0.0_{0.0}$ \\
	Importance & $3.9_{0.6}$ & $3.9_{0.9}$ & $3.9_{0.3}$ & $0.0_{0.0}$ \\
	Number & $2.7_{0.5}$ & $3.7_{0.1}$ & $2.3_{0.1}$ & $0.0_{0.0}$ \\
	\makecell[l]{Number+\\Base Type} & $3.5_{0.2}$ & $5.5_{0.3}$ & $3.8_{0.1}$ & $0.0_{0.0}$ \\
	\makecell[l]{Number+\\Abstract Type} & $3.6_{0.4}$ & $2.8_{1.0}$ & $3.3_{0.4}$ & $0.0_{0.1}$ \\
	\midrule
	Macro Avg. & $3.3_{0.2}$ & $3.3_{0.4}$ & $2.1_{0.0}$ & $0.0_{0.0}$ \\
    \bottomrule
    \end{tabular}}
    \caption{
        Analysis of JSON format correctness on all instruction categories.
        We report the JSON decoding error rates (\%) in this table.
        \DevEx
        }
    \label{tab:analysis-json-format-correctness}
\end{table}

\begin{table}[ht]
    \centering
    \small
    \setlength{\tabcolsep}{0.5mm}{
    \begin{tabular}{l|cccc}
    \toprule
    Instruction & ChatGPT & \makecell[c]{ChatGPT\\w/Demo} & \modelname-1b & \modelname-7b \\
    \midrule
	Number & $26.8_{0.4}$ & $13.4_{0.5}$ & $6.5_{0.0}$ & $0.6_{0.0}$ \\
	Importance & $25.6_{1.1}$ & $18.7_{0.2}$ & $7.9_{0.3}$ & $0.8_{0.0}$ \\
	\makecell[l]{Number+\\Base Type} & $43.4_{0.4}$ & $37.3_{0.0}$ & $38.0_{0.9}$ & $29.0_{0.2}$ \\
	\makecell[l]{Number+\\Abstract Type} & $48.3_{0.9}$ & $40.8_{0.4}$ & $42.3_{0.8}$ & $37.3_{0.1}$ \\
	\midrule
	Macro Avg. & $36.0_{0.1}$ & $27.5_{0.0}$ & $23.7_{0.1}$ & $16.9_{0.0}$ \\
    \bottomrule
    \end{tabular}}
    \caption{
        Analysis of following entity number constraints.
        We report error rates (\%) of predictions that do not have the same number of entities as the instruction.
        \DevEx
    }
    \label{tab:analysis-entity-number-constraint}
\end{table}

\begin{table}[ht]
    \centering
    \small
    \setlength{\tabcolsep}{0.5mm}{
    \begin{tabular}{l|cccc}
    \toprule
    Instruction & ChatGPT & \makecell[c]{ChatGPT\\w/Demo} & \modelname-1b & \modelname-7b \\
    \midrule
	Base Type & $72.4_{0.8}$ & $48.8_{0.1}$ & $38.1_{0.3}$ & $22.4_{0.6}$ \\
	Abstract Type & $65.1_{2.1}$ & $50.7_{0.7}$ & $16.9_{0.3}$ & $4.1_{0.2}$ \\
	\makecell[l]{Number+\\Base Type} & $65.8_{0.5}$ & $47.3_{0.4}$ & $42.5_{0.8}$ & $24.7_{0.3}$ \\
	\makecell[l]{Number+\\Abstract Type} & $57.1_{1.6}$ & $43.5_{0.8}$ & $31.1_{0.7}$ & $25.5_{1.5}$ \\
	\midrule
	Macro Avg. & $65.1_{0.8}$ & $47.6_{0.1}$ & $32.2_{0.5}$ & $19.2_{0.5}$ \\
    \bottomrule
    \end{tabular}}
    \caption{
        Analysis of following entity type constraints.
        We report error rates (\%) of predictions that only have entities in specific types from instructions.
        \DevEx
    }
    \label{tab:analysis-type-constraint}
\end{table}

\begin{table*}[t]
    \centering
    \small
    \setlength{\tabcolsep}{1mm}{
    \begin{tabularx}{\linewidth}{l
    >{\raggedright\arraybackslash}X
    >{\raggedright\arraybackslash}X
    c}
    \toprule
    Categories & Manually Designed Templates & Rephrased Templates & \makecell[c]{\#Rephrased\\Templates} \\
    \midrule
    Default & Extract entities. & Identify the entities present in the text. & 219 \\
    Base Type & Extract entities in types \{types\}. & Please identify the entities falling under the categories \{types\}. & 48\\
    Abstract Type & Extract entities in types \{types\}. & Please identify the entities falling under the categories \{types\}. & 48\\
    Description & Extract entities with following descriptions: \{descriptions\}. & Can you identify the entities described as followed: \{descriptions\}? & 104 \\
    Importance & Extract the most important \{num\} entities. & Retrieve the \{number\} most essential entities. & 62 \\
    Number & Extract \{num\} entities. & Fetch \{number\} entities. & 49 \\
    Number+Base Type & Extract \{num\} entities in types \{types\}. & Could you identify \{number\} entities belonging to \{types\}? & 117 \\
    Number+Abstract Type & Extract \{num\} entities in types \{types\}. & Retrieve \{number\} entities belonging to \{types\}. & 117 \\
    \bottomrule
    \end{tabularx}}
    \caption{
        Details of Instructions.
        \{num\}, \{types\}, \{descriptions\} are placeholders for entity numbers, types, and descriptions.
        We only display plural forms of templates in this table, but the templates will be different for singular or plural.
        We only show one example of rephrased templates for each category.
    }
    \label{tab:instruction}
\end{table*}

\begin{table*}[t]
    \centering
    \small
    \setlength{\tabcolsep}{1mm}{
    \begin{tabular}{c|ccc|cccc|cccc}
    \toprule
    \multirow{2}{*}{Split} & \multicolumn{3}{c}{Corpus}\vline & \multicolumn{4}{c}{Ontology}\vline & \multicolumn{3}{c}{Entity Info Density} \\
    & \#Article & \#Mention & \#Triplets & \#Ent. & \#Aliases & \#Rel. & \#Types & \%Desc. & \%Aliases & \%Types \\
    \midrule
    Train & 11,447,454 & 39,930,663 & 19,184,948 & 2,234,052 & 840,401 & 962 & 21,350 & 93.5 & 64.2 & 71.5 \\
    \midrule
     Dev & 2,710 & \makecell[c]{13,601\\unseen:3038} & 5,915 & \makecell[c]{6,868\\unseen:1417} & 8,812 & 234 & %
     1,163
     & 94.6 & 55.6 & 70.2 \\
    \midrule
    Test & 24,393 & \makecell[c]{109,411\\unseen:29,612} & 45,758 & \makecell[c]{39,086\\unseen:9,474} & 37,809 & 398 & %
    3,306 & 92.7 & 52.6 & 70.7 \\
    \bottomrule
    \end{tabular}}
    \caption{
        Statistics of the instruction-following open-world dataset \datasetname.
        We report the statistics of the corpus, including the number of articles, mentions, and triplets in the left section.
        The ontology statistics, including the number of unique entities, aliases, relations, and types, are reported in the middle section.
        The right section shows the proportions of mentions with descriptions, aliases, and types, respectively.
        Ent., Rel., and Desc. are short for entities, relations, and descriptions.
    }
    \label{tab:data-statistics}
\end{table*}

\begin{table*}[t]
    \centering
    \small
    \setlength{\tabcolsep}{1mm}{
    \begin{tabular}{c|ccccccccc}
    \toprule
    \multirow{2}{*}{\textbf{Method}} & \multirow{2}{*}{\textbf{Partition}} & \multirow{2}{*}{\textbf{Instruction}} &\textbf{MD} & \multicolumn{2}{c}{\textbf{EL}} & \textbf{ET} & \textbf{OpenRE} & \textbf{EIG~(Desc.)} & \textbf{EIG~(Aliases)} \\
    & & & F1 & F1(T=1) & F1(T=0.8) & F1 & F1(CaRB) & F1(ROUGE-L) & F1 \\
    \midrule
	\multirow{8}{*}{ChatGPT} & \multirow{4}{*}{unseen} & Base Type & $56.3_{1.0}$ & $31.4_{1.9}$ & $34.5_{2.3}$ & $16.5_{0.5}$ & $23.3_{0.9}$ & $41.4_{0.5}$ & $17.9_{1.2}$ \\
	 & & Abstract Type & $47.2_{0.6}$ & $23.7_{0.9}$ & $24.6_{0.9}$ & $10.3_{0.4}$ & $23.0_{1.6}$ & $39.5_{2.1}$ & $19.7_{0.4}$ \\
	 & & Description & $51.8_{0.1}$ & $24.5_{0.4}$ & $25.3_{0.3}$ & $3.3_{0.1}$ & $22.0_{0.0}$ & $56.3_{0.2}$ & $14.7_{0.6}$ \\
     \cline{2-10}
	 & & Macro Avg. & $51.8_{0.5}$ & $26.5_{0.5}$ & $28.1_{0.8}$ & $10.0_{0.3}$ & $22.8_{0.6}$ & $45.8_{0.6}$ & $17.4_{0.4}$ \\
     \cline{2-10}
	 & \multirow{4}{*}{seen} & Base Type & $50.1_{0.5}$ & $26.4_{0.9}$ & $27.4_{0.9}$ & $15.3_{0.6}$ & $21.0_{0.8}$ & $37.8_{1.1}$ & $13.6_{0.6}$ \\
	 & & Abstract Type & $47.5_{0.2}$ & $18.1_{0.8}$ & $19.3_{1.0}$ & $7.6_{0.5}$ & $21.0_{0.0}$ & $36.6_{0.8}$ & $16.7_{1.4}$ \\
	 & & Description & $46.6_{0.8}$ & $21.6_{1.0}$ & $22.9_{1.2}$ & $1.7_{0.3}$ & $22.0_{0.8}$ & $45.1_{1.3}$ & $25.0_{3.8}$ \\
    \cline{2-10}
	 & & Macro Avg. & $48.1_{0.2}$ & $22.1_{0.7}$ & $23.2_{0.8}$ & $8.2_{0.4}$ & $21.3_{0.5}$ & $39.8_{1.0}$ & $18.4_{1.6}$ \\
     \midrule
	\multirow{8}{*}{ChatGPT w/Demo} & \multirow{4}{*}{unseen} & Base Type & $58.4_{0.7}$ & $42.0_{0.4}$ & $45.5_{0.3}$ & $30.0_{0.2}$ & $24.0_{0.0}$ & $45.8_{0.6}$ & $17.4_{1.4}$ \\
	 & & Abstract Type & $47.3_{0.0}$ & $37.6_{0.8}$ & $39.2_{0.9}$ & $16.8_{0.4}$ & $21.5_{1.5}$ & $51.9_{1.4}$ & $25.0_{2.7}$ \\
	 & & Description & $52.2_{0.1}$ & $39.2_{0.2}$ & $40.0_{0.3}$ & $6.7_{0.1}$ & $17.5_{0.5}$ & $83.5_{0.3}$ & $24.4_{0.5}$ \\
     \cline{2-10}
	 & & Macro Avg. & $52.6_{0.2}$ & $39.6_{0.3}$ & $41.6_{0.3}$ & $17.8_{0.2}$ & $21.0_{0.3}$ & $60.4_{0.8}$ & $22.3_{0.3}$ \\
     \cline{2-10}
	 & \multirow{4}{*}{seen} & Base Type & $55.1_{0.4}$ & $40.5_{0.6}$ & $41.5_{0.5}$ & $32.9_{0.3}$ & $21.0_{0.0}$ & $47.7_{0.1}$ & $12.6_{0.6}$ \\
	 & & Abstract Type & $49.1_{0.3}$ & $37.0_{0.2}$ & $38.4_{0.2}$ & $17.2_{0.2}$ & $21.0_{1.0}$ & $52.6_{0.6}$ & $18.7_{0.3}$ \\
	 & & Description & $47.6_{0.8}$ & $34.1_{1.4}$ & $35.4_{1.5}$ & $4.6_{0.1}$ & $17.0_{0.0}$ & $78.6_{0.3}$ & $38.7_{1.2}$ \\
     \cline{2-10}
	 & & Macro Avg. & $50.6_{0.2}$ & $37.2_{0.3}$ & $38.4_{0.4}$ & $18.2_{0.2}$ & $19.7_{0.3}$ & $59.7_{0.2}$ & $23.4_{0.7}$ \\
     \midrule
	\multirow{8}{*}{PIVOINE-1b} & \multirow{4}{*}{unseen} & Base Type & $62.7_{0.5}$ & $51.3_{0.4}$ & $52.8_{0.4}$ & $38.3_{0.1}$ & $62.7_{0.5}$ & $64.7_{0.5}$ & $74.1_{0.1}$ \\
	 & & Abstract Type & $54.0_{0.2}$ & $44.6_{0.5}$ & $45.2_{0.5}$ & $32.2_{0.5}$ & $53.0_{1.4}$ & $68.7_{1.3}$ & $70.0_{1.0}$ \\
	 & & Description & $62.6_{0.4}$ & $50.6_{0.3}$ & $51.4_{0.3}$ & $44.6_{0.2}$ & $56.7_{0.5}$ & $75.1_{0.1}$ & $70.3_{0.1}$ \\
     \cline{2-10}
	 & & Macro Avg. & $59.8_{0.3}$ & $48.8_{0.2}$ & $49.8_{0.3}$ & $38.3_{0.2}$ & $57.4_{0.6}$ & $69.5_{0.5}$ & $71.5_{0.3}$ \\
     \cline{2-10}
	 & \multirow{4}{*}{seen} & Base Type & $61.7_{0.2}$ & $50.1_{0.1}$ & $51.0_{0.2}$ & $44.0_{0.1}$ & $56.3_{0.5}$ & $64.2_{0.1}$ & $70.9_{0.2}$ \\
	 & & Abstract Type & $59.0_{0.1}$ & $48.5_{0.2}$ & $49.3_{0.2}$ & $40.8_{0.1}$ & $56.0_{0.8}$ & $62.0_{0.2}$ & $70.1_{0.2}$ \\
	 & & Description & $67.8_{0.8}$ & $55.3_{0.6}$ & $57.1_{0.8}$ & $42.3_{0.2}$ & $62.7_{1.7}$ & $90.6_{0.0}$ & $75.0_{0.7}$ \\
     \cline{2-10}
	 & & Macro Avg. & $62.8_{0.3}$ & $51.3_{0.2}$ & $52.4_{0.3}$ & $42.4_{0.1}$ & $58.3_{0.8}$ & $72.3_{0.0}$ & $72.0_{0.3}$ \\
     \midrule
	\multirow{8}{*}{PIVOINE-7b} & \multirow{4}{*}{unseen} & Base Type & $84.1_{0.4}$ & $74.4_{0.5}$ & $75.6_{0.6}$ & $56.0_{0.3}$ & $69.0_{0.8}$ & $74.1_{0.2}$ & $76.7_{0.1}$ \\
	 & & Abstract Type & $74.5_{0.2}$ & $65.0_{0.2}$ & $65.5_{0.1}$ & $52.0_{0.3}$ & $66.3_{0.5}$ & $78.8_{0.1}$ & $80.6_{0.1}$ \\
	 & & Description & $88.3_{0.1}$ & $75.9_{0.1}$ & $77.0_{0.1}$ & $64.0_{0.1}$ & $72.3_{0.5}$ & $87.0_{0.0}$ & $79.4_{0.1}$ \\
     \cline{2-10}
	 & & Macro Avg. & $82.3_{0.2}$ & $71.7_{0.2}$ & $72.7_{0.2}$ & $57.3_{0.2}$ & $69.2_{0.2}$ & $80.0_{0.0}$ & $78.9_{0.1}$ \\
     \cline{2-10}
	 & \multirow{4}{*}{seen} & Base Type & $76.3_{0.2}$ & $66.7_{0.1}$ & $67.5_{0.1}$ & $59.2_{0.2}$ & $65.7_{0.9}$ & $76.9_{0.2}$ & $78.6_{0.0}$ \\
	 & & Abstract Type & $78.5_{0.3}$ & $69.8_{0.2}$ & $70.4_{0.2}$ & $59.6_{0.1}$ & $69.7_{0.5}$ & $76.8_{0.1}$ & $80.0_{0.1}$ \\
	 & & Description & $90.5_{0.1}$ & $80.9_{0.1}$ & $81.4_{0.1}$ & $59.6_{0.3}$ & $80.0_{0.0}$ & $97.0_{0.0}$ & $81.3_{0.0}$ \\
     \cline{2-10}
	 & & Macro Avg. & $81.8_{0.1}$ & $72.5_{0.1}$ & $73.1_{0.1}$ & $59.5_{0.1}$ & $71.8_{0.4}$ & $83.6_{0.1}$ & $80.0_{0.0}$ \\
    \bottomrule
    \end{tabular}}
    \caption{Main results of the instruction generalization evaluation on three instruction categories ``Base Type'', ``Abstract Type'', and ``Description''.
    Headers are the same as \Cref{tab:main-results-overall-performance}.
    Partition denotes the unseen and seen instructions.
    \DevEx
    }
    \label{tab:main-results-instruction-generalization}
\end{table*}

\begin{table*}[t]
    \centering
    \small
    \setlength{\tabcolsep}{0.4mm}{
    \begin{tabular}{lccccccccccccccc}
    \toprule
    \multirow{2}{*}{Instruction} & \multicolumn{3}{c}{GENRE} & \multicolumn{3}{c}{ChatGPT}  & \multicolumn{3}{c}{ChatGPT w/ Demo} & \multicolumn{3}{c}{\modelname-1b} & \multicolumn{3}{c}{\modelname-7b} \\
    & P & R & F1 & P & R & F1 & P & R & F1 & P & R & F1 & P & R & F1 \\
    \midrule
	Default & $38.8$ & $50.0$ & $43.7$ & $53.4_{0.1}$ & $57.5_{0.2}$ & $55.4_{0.1}$ & $51.2_{0.1}$ & $51.9_{0.1}$ & $51.6_{0.1}$ & $71.1_{0.0}$ & $57.3_{0.0}$ & $63.5_{0.0}$ & $82.0_{0.0}$ & $76.4_{0.0}$ & $79.1_{0.0}$ \\
	Base Type & $--$ & $--$ & $--$ & $41.9_{0.4}$ & $66.3_{1.0}$ & $51.4_{0.6}$ & $47.9_{0.1}$ & $66.8_{0.3}$ & $55.8_{0.2}$ & $65.8_{0.1}$ & $58.5_{0.0}$ & $61.9_{0.1}$ & $76.0_{0.2}$ & $80.2_{0.1}$ & $78.1_{0.1}$ \\
	Abstract Type & $--$ & $--$ & $--$ & $38.2_{0.4}$ & $62.5_{0.2}$ & $47.4_{0.3}$ & $42.5_{0.3}$ & $57.1_{0.1}$ & $48.7_{0.2}$ & $61.2_{0.1}$ & $54.9_{0.2}$ & $57.9_{0.1}$ & $76.8_{0.1}$ & $78.3_{0.2}$ & $77.5_{0.2}$ \\
	Description & $--$ & $--$ & $--$ & $43.9_{0.2}$ & $61.1_{0.4}$ & $51.1_{0.0}$ & $44.7_{0.0}$ & $60.9_{0.2}$ & $51.6_{0.1}$ & $65.9_{0.1}$ & $60.8_{0.6}$ & $63.2_{0.3}$ & $88.7_{0.1}$ & $88.4_{0.1}$ & $88.5_{0.1}$ \\
	Importance & $--$ & $--$ & $--$ & $46.8_{0.5}$ & $47.9_{0.2}$ & $47.4_{0.3}$ & $47.5_{0.4}$ & $48.2_{0.1}$ & $47.8_{0.1}$ & $64.1_{0.1}$ & $57.6_{0.5}$ & $60.7_{0.3}$ & $74.8_{0.2}$ & $74.7_{0.1}$ & $74.7_{0.1}$ \\
    \midrule
	Macro Avg. & $--$ & $--$ & $--$ & $44.9_{0.2}$ & $59.1_{0.1}$ & $50.5_{0.2}$ & $46.8_{0.1}$ & $57.0_{0.0}$ & $51.1_{0.1}$ & $65.6_{0.0}$ & $57.8_{0.0}$ & $61.4_{0.0}$ & $79.7_{0.1}$ & $79.6_{0.0}$ & $79.6_{0.0}$ \\
    \bottomrule
    \end{tabular}}
    
    \caption{
        Main results of mention detection~(MD).
        We report precision, recall, and F1 score of each model in this table.
        The subscript scores are the deviation under three different rephrased instructions for each instruction category.
        \DevEx
        \textsc{GENRE} is an closed-world generative IE baseline without the instruction-following ability, so we only report GENRE scores with the default instruction.
    }
    \label{tab:main-results-md}
\end{table*}

\begin{table*}[t]
    \centering
    \small
    \setlength{\tabcolsep}{1mm}{
    \begin{tabular}{lcccccccccc}
    \toprule
    \multirow{2}{*}{Instruction} & \multicolumn{2}{c}{GENRE} & \multicolumn{2}{c}{ChatGPT}  & \multicolumn{2}{c}{ChatGPT w/Demo} & \multicolumn{2}{c}{\modelname-1b} & \multicolumn{2}{c}{\modelname-7b} \\
    & F1(T=1) & F1(T=0.8) & F1(T=1) & F1(T=0.8) & F1(R=1) & F1(R=0.8) & F1(R=1) & F1(R=0.8) & F1(R=1) & F1(R=0.8)\\
    \midrule
	Default & 17.2 & 20.1 & $27.4_{0.1}$ & $28.3_{0.0}$ & $39.6_{0.0}$ & $41.0_{0.0}$ & $50.3_{0.0}$ & $51.3_{0.0}$ & $69.4_{0.1}$ & $70.1_{0.0}$ \\
	Base Type & $--$ & $--$ & $27.4_{0.4}$ & $28.8_{0.4}$ & $40.8_{0.4}$ & $42.3_{0.4}$ & $50.4_{0.0}$ & $51.4_{0.1}$ & $68.5_{0.1}$ & $69.3_{0.1}$ \\
	Abstract Type & $--$ & $--$ & $19.3_{0.5}$ & $20.4_{0.6}$ & $37.1_{0.3}$ & $38.6_{0.3}$ & $47.6_{0.2}$ & $48.4_{0.2}$ & $68.7_{0.2}$ & $69.2_{0.2}$ \\
	Description & $--$ & $--$ & $24.2_{0.3}$ & $24.9_{0.1}$ & $38.5_{0.0}$ & $39.4_{0.0}$ & $51.2_{0.2}$ & $52.1_{0.2}$ & $76.4_{0.1}$ & $77.5_{0.1}$ \\
	Importance & $--$ & $--$ & $26.7_{0.4}$ & $28.0_{0.4}$ & $36.8_{0.1}$ & $38.2_{0.0}$ & $48.7_{0.2}$ & $49.8_{0.3}$ & $66.2_{0.1}$ & $67.2_{0.1}$ \\
    \midrule
	Macro Avg. & $--$ & $--$ & $25.0_{0.2}$ & $26.1_{0.2}$ & $38.6_{0.0}$ & $39.9_{0.0}$ & $49.6_{0.0}$ & $50.6_{0.0}$ & $69.8_{0.1}$ & $70.7_{0.1}$ \\
    \bottomrule
    \end{tabular}}
    \caption{
        Main results of end-to-end entity linking~(EL).
        We report F1 score calculated by title matching with ROUGE-L F1 threshold 1 and 0.8 in this table.
        \DevEx
        \textsc{GENRE} is an closed-world generative IE baseline without the instruction-following ability, so we only report GENRE scores with the default instruction.
    }
    \label{tab:main-results-end2end-el}
\end{table*}

\begin{table*}[t]
    \centering
    \small
    \setlength{\tabcolsep}{3.5mm}{
    \begin{tabular}{llccccc}
    \toprule
    Partition & Instruction & GENRE & ChatGPT & ChatGPT w/Demo & \modelname-1 & \modelname-7b \\
    \midrule
	 \multirow{6}{*}{\makecell[c]{Before\\05/30/2022}} & Default & $55.3$ & $56.5_{0.2}$ & $50.1_{0.2}$ & $56.8_{0.0}$ & $77.8_{0.0}$ \\
	 & Base Type & $--$ & $66.1_{1.2}$ & $65.4_{0.5}$ & $59.4_{0.2}$ & $86.5_{0.1}$ \\
	 & Abstract Type & $--$ & $63.6_{0.7}$ & $56.7_{0.1}$ & $56.4_{0.1}$ & $82.9_{0.1}$ \\
	 & Description & $--$ & $60.9_{0.3}$ & $61.1_{0.2}$ & $62.8_{0.8}$ & $90.4_{0.1}$ \\
	 & Importance & $--$ & $49.1_{0.1}$ & $49.0_{0.1}$ & $60.7_{0.6}$ & $80.7_{0.1}$ \\
	\cline{2-7}
	& Macro Avg. & $--$ & $59.3_{0.3}$ & $56.5_{0.0}$ & $59.2_{0.1}$ & $83.7_{0.0}$ \\
	\midrule
	 \multirow{6}{*}{\makecell[c]{After\\05/30/2022}} & Default & $33.6$ & $60.5_{0.1}$ & $57.4_{0.2}$ & $58.7_{0.0}$ & $71.8_{0.0}$ \\
	 & Base Type & $--$ & $66.7_{1.1}$ & $69.0_{0.0}$ & $57.1_{0.2}$ & $70.4_{0.4}$ \\
	 & Abstract Type & $--$ & $60.6_{0.8}$ & $57.8_{0.0}$ & $52.5_{0.4}$ & $70.9_{0.5}$ \\
	 & Description & $--$ & $61.5_{0.6}$ & $60.5_{0.2}$ & $56.5_{0.2}$ & $84.1_{0.2}$ \\
	 & Importance & $--$ & $45.3_{0.5}$ & $46.4_{0.1}$ & $50.9_{0.5}$ & $61.6_{0.3}$ \\
	\cline{2-7}
	& Macro Avg. & $--$ & $58.9_{0.2}$ & $58.2_{0.0}$ & $55.1_{0.0}$ & $71.7_{0.1}$ \\
    \bottomrule
    \end{tabular}}
    \caption{
        Results of generalization study on mention detection~(MD).
        We report recalls of mention spans in this table.
        \PartitionEx
        \DevEx
        \textsc{GENRE} is an closed-world generative IE baseline without the instruction-following ability, so we only report GENRE scores with the default instruction.
    }
    \label{tab:main-results-md-generalization}
\end{table*}

\begin{table*}[t]
    \centering
    \small
    \setlength{\tabcolsep}{0.8mm}{
    \begin{tabular}{llcccccccccc}
    \toprule
    \multirow{2}{*}{Partition}
    & \multirow{2}{*}{Instruction} & \multicolumn{2}{c}{GENRE} & \multicolumn{2}{c}{ChatGPT} & \multicolumn{2}{c}{ChatGPT w/Demo} & \multicolumn{2}{c}{\modelname-1} & \multicolumn{2}{c}{\modelname-7b} \\
    & & R(T=1) & R(T=0.8) & R(T=1) & R(T=0.8) & R(T=1) & R(T=0.8) & R(T=1) & R(T=0.8)& R(T=1) & R(T=0.8)\\
    \midrule
    \multirow{6}{*}{\makecell[c]{Before\\05/30/2022}} & Default & $23.8$ & $24.7$ & $29.6_{0.1}$ & $30.2_{0.1}$ & $40.2_{0.1}$ & $41.0_{0.1}$ & $48.5_{0.0}$ & $49.0_{0.0}$ & $73.0_{0.0}$ & $73.3_{0.0}$ \\
	 & Base Type & $--$ & $--$ & $39.3_{0.5}$ & $39.7_{0.4}$ & $50.4_{0.9}$ & $51.0_{0.9}$ & $52.9_{0.0}$ & $53.0_{0.0}$ & $81.6_{0.1}$ & $81.8_{0.1}$ \\
	 & Abstract Type & $--$ & $--$ & $29.5_{0.9}$ & $29.9_{0.9}$ & $47.8_{0.2}$ & $48.3_{0.2}$ & $49.5_{0.1}$ & $49.8_{0.1}$ & $78.0_{0.3}$ & $78.1_{0.3}$ \\
	 & Description & $--$ & $--$ & $30.9_{0.7}$ & $31.2_{0.7}$ & $47.7_{0.1}$ & $48.3_{0.1}$ & $55.5_{0.6}$ & $55.8_{0.6}$ & $85.3_{0.2}$ & $85.7_{0.2}$ \\
	 & Importance & $--$ & $--$ & $29.8_{0.6}$ & $30.4_{0.6}$ & $39.1_{0.2}$ & $39.9_{0.1}$ & $52.7_{0.5}$ & $53.3_{0.5}$ & $76.3_{0.0}$ & $76.6_{0.0}$ \\
	\cline{2-12}
	& Macro Avg. & $--$ & $--$ & $31.8_{0.2}$ & $32.3_{0.2}$ & $45.0_{0.1}$ & $45.7_{0.1}$ & $51.8_{0.0}$ & $52.2_{0.0}$ & $78.8_{0.1}$ & $79.1_{0.1}$ \\
	\midrule
	 \multirow{6}{*}{\makecell[c]{After\\05/30/2022}} & Default & $0$ & $3.6$ & $24.7_{0.1}$ & $27.1_{0.1}$ & $38.7_{0.2}$ & $41.8_{0.2}$ & $35.4_{0.0}$ & $37.8_{0.0}$ & $48.3_{0.1}$ & $50.4_{0.1}$ \\
	 & Base Type & $--$ & $--$ & $29.3_{0.9}$ & $33.1_{1.1}$ & $46.5_{0.1}$ & $50.0_{0.2}$ & $39.2_{0.1}$ & $41.4_{0.1}$ & $52.5_{0.6}$ & $54.6_{0.7}$ \\
	 & Abstract Type & $--$ & $--$ & $18.9_{1.0}$ & $21.9_{1.0}$ & $36.8_{0.2}$ & $40.3_{0.3}$ & $38.3_{0.4}$ & $39.7_{0.4}$ & $55.5_{0.4}$ & $56.8_{0.4}$ \\
	 & Description & $--$ & $--$ & $24.8_{0.9}$ & $27.0_{0.9}$ & $40.7_{0.0}$ & $42.9_{0.0}$ & $36.4_{0.2}$ & $38.3_{0.2}$ & $57.8_{0.1}$ & $60.3_{0.1}$ \\
	 & Importance & $--$ & $--$ & $21.3_{0.1}$ & $23.6_{0.3}$ & $32.8_{0.4}$ & $35.4_{0.3}$ & $32.3_{0.2}$ & $34.3_{0.3}$ & $44.5_{0.3}$ & $46.8_{0.2}$ \\
	\cline{2-12}
	& Macro Avg. & $--$ & $--$ & $23.8_{0.2}$ & $26.5_{0.3}$ & $39.1_{0.0}$ & $42.1_{0.0}$ & $36.3_{0.1}$ & $38.3_{0.1}$ & $51.7_{0.2}$ & $53.8_{0.2}$ \\
    \bottomrule
    \end{tabular}}
    \caption{
        Results of generalization study on end-to-end entity linking~(EL).
        We report recalls calculated by title matching with ROUGE-L F1 threshold 1 and 0.8 in this table.
        \PartitionEx
        \DevEx
        \textsc{GENRE} is an closed-world generative IE baseline without the instruction-following ability, so we only report GENRE scores with the default instruction.
    }
    \label{tab:main-results-ie-generalization}
\end{table*}

\begin{table*}[t]
    \centering
    \small
    \setlength{\tabcolsep}{0.3mm}{
    \begin{tabular}{l|ccc|ccc|ccc|ccc|ccc}
    \toprule
    \multirow{2}{*}{Instruction} & \multicolumn{3}{c}{OpenIE6}\vline & \multicolumn{3}{c}{ChatGPT}\vline & \multicolumn{3}{c}{ChatGPT w/Demo}\vline & \multicolumn{3}{c}{\modelname-1b}\vline & \multicolumn{3}{c}{\modelname-7b} \\
    & P & R & F1 & P & R & F1 & P & R & F1 & P & R & F1 & P & R & F1 \\
    \midrule
	Default & $13.0$ & $18.5$ & $15.2$ & $17.8_{0.0}$ & $36.9_{0.1}$ & $23.7_{0.5}$ & $16.4_{0.0}$ & $37.8_{0.2}$ & $22.0_{0.0}$ & $54.1_{0.0}$ & $55.8_{0.0}$ & $54.0_{0.0}$ & $65.7_{0.1}$ & $68.1_{0.0}$ & $66.3_{0.5}$ \\
	Base Type & $--$ & $--$ & $--$ & $16.1_{0.3}$ & $34.6_{1.3}$ & $21.7_{0.5}$ & $15.2_{0.1}$ & $43.0_{0.3}$ & $22.0_{0.0}$ & $56.6_{0.7}$ & $61.6_{0.5}$ & $58.3_{0.5}$ & $60.3_{1.2}$ & $75.7_{0.4}$ & $66.7_{0.9}$ \\
	Abstract Type & $--$ & $--$ & $--$ & $15.7_{0.6}$ & $37.8_{0.3}$ & $22.0_{0.8}$ & $14.6_{0.4}$ & $46.4_{0.6}$ & $21.5_{0.5}$ & $56.7_{1.0}$ & $54.5_{0.1}$ & $55.0_{0.8}$ & $66.4_{0.3}$ & $71.5_{0.1}$ & $68.3_{0.5}$ \\
	Description & $--$ & $--$ & $--$ & $16.6_{0.1}$ & $35.8_{0.3}$ & $22.0_{0.0}$ & $12.0_{0.2}$ & $35.2_{0.8}$ & $17.5_{0.5}$ & $52.3_{0.6}$ & $63.8_{0.6}$ & $57.0_{0.0}$ & $69.3_{0.4}$ & $77.7_{0.1}$ & $73.0_{0.0}$ \\
	Importance & $--$ & $--$ & $--$ & $18.1_{0.2}$ & $35.8_{0.4}$ & $23.7_{0.5}$ & $17.2_{0.3}$ & $39.6_{0.1}$ & $23.5_{0.5}$ & $54.9_{0.3}$ & $57.9_{0.4}$ & $55.7_{0.5}$ & $62.4_{0.1}$ & $68.8_{0.2}$ & $65.0_{0.0}$ \\
    \midrule
	Macro Avg. & $--$ & $--$ & $--$ & $16.9_{0.2}$ & $36.2_{0.2}$ & $22.6_{0.3}$ & $15.1_{0.1}$ & $40.4_{0.2}$ & $21.3_{0.1}$ & $54.9_{0.4}$ & $58.7_{0.2}$ & $56.0_{0.3}$ & $64.8_{0.3}$ & $72.4_{0.0}$ & $67.9_{0.2}$ \\
    \bottomrule
    \end{tabular}}
    \caption{
    Main results of open relation extraction~(Open RE).
    We report precision, recall, and F1 with the CaRB scoring based on the ROUGE-L matcher.
    \DevEx
    \textsc{OpenIE6} is an openIE baseline without the instruction-following ability, so we only report OpenIE6 scores with the default instruction.
    }
    \label{tab:main-results-openre-generalization}
\end{table*}

\begin{table*}[t]
    \centering
    \small
    \setlength{\tabcolsep}{1mm}{
    \begin{tabular}{ll|ccccc}
    \toprule
    Partition & Instruction & OpenIE6 & ChatGPT & ChatGPT w/ Demo & \modelname-1b & \modelname-7b \\
    \midrule
	 \multirow{6}{*}{\makecell[c]{Before\\05/30/2022}} & Default & $21.9$ & $35.1_{0.1}$ & $35.9_{0.1}$ & $57.7_{0.0}$ & $73.3_{0.0}$ \\
	 & Base Type & $--$ & $32.1_{1.7}$ & $40.4_{0.8}$ & $63.6_{0.6}$ & $83.7_{0.5}$ \\
	 & Abstract Type & $--$ & $36.0_{0.6}$ & $43.7_{0.6}$ & $58.8_{0.1}$ & $78.4_{0.1}$ \\
	 & Description & $--$ & $33.3_{0.5}$ & $33.0_{0.7}$ & $67.9_{0.6}$ & $84.1_{0.1}$ \\
	 & Importance & $--$ & $34.2_{0.5}$ & $38.5_{0.2}$ & $61.0_{0.6}$ & $77.4_{0.2}$ \\
	\cline{2-7}
	& Macro Avg. & $--$ & $34.1_{0.2}$ & $38.3_{0.0}$ & $61.8_{0.3}$ & $79.4_{0.1}$ \\
	\midrule
	 \multirow{6}{*}{\makecell[c]{After\\05/30/2022}} & Default & $18.2$ & $32.7_{0.2}$ & $34.0_{0.2}$ & $35.8_{0.1}$ & $36.1_{0.1}$ \\
	 & Base Type & $--$ & $23.6_{0.8}$ & $27.8_{0.6}$ & $39.5_{0.2}$ & $37.9_{0.2}$ \\
	 & Abstract Type & $--$ & $28.0_{1.1}$ & $33.5_{1.3}$ & $32.6_{0.0}$ & $36.4_{0.4}$ \\
	 & Description & $--$ & $24.8_{0.2}$ & $22.4_{0.5}$ & $30.3_{0.4}$ & $35.3_{0.2}$ \\
	 & Importance & $--$ & $26.2_{0.7}$ & $29.0_{0.6}$ & $34.1_{0.5}$ & $34.6_{0.1}$ \\
	\cline{2-7}
	& Macro Avg. & $--$ & $27.1_{0.5}$ & $29.3_{0.5}$ & $34.5_{0.1}$ & $36.1_{0.1}$ \\
    \bottomrule
    \end{tabular}}
    \caption{
        Results of generalization study on open relation extraction~(Open RE).
        We report recall with the CaRB scoring based on the ROUGE-L matcher.
        \PartitionEx
        \DevEx
        \textsc{OpenIE6} is an openIE baseline without the instruction-following ability, so we only report OpenIE6 scores with the default instruction.
    }
    \label{tab:main-results-openre-generalization}
\end{table*}

\begin{table*}[t]
    \centering
    \small
    \setlength{\tabcolsep}{0.4mm}{
    \begin{tabular}{lcccccccccccc}
    \toprule
    \multirow{2}{*}{Instruction} & \multicolumn{3}{c}{ChatGPT}  & \multicolumn{3}{c}{ChatGPT w/ Demo} & \multicolumn{3}{c}{\modelname-1b} & \multicolumn{3}{c}{\modelname-7b} \\
    & P & R & F1 & P & R & F1 & P & R & F1 & P & R & F1 \\
    \midrule
	Default & $2.7_{0.0}$ & $3.8_{0.0}$ & $3.1_{0.0}$ & $6.4_{0.0}$ & $7.5_{0.0}$ & $6.9_{0.0}$ & $46.3_{0.1}$ & $37.9_{0.1}$ & $41.7_{0.1}$ & $61.3_{0.0}$ & $55.3_{0.0}$ & $58.1_{0.0}$ \\
	Base Type & $13.0_{0.3}$ & $19.4_{0.5}$ & $15.6_{0.4}$ & $29.9_{0.2}$ & $35.0_{0.4}$ & $32.3_{0.3}$ & $46.0_{0.1}$ & $39.9_{0.2}$ & $42.7_{0.1}$ & $59.3_{0.2}$ & $57.6_{0.1}$ & $58.4_{0.1}$ \\
	Abstract Type & $8.9_{0.5}$ & $7.5_{0.2}$ & $8.2_{0.3}$ & $16.2_{0.3}$ & $18.2_{0.2}$ & $17.1_{0.3}$ & $42.7_{0.2}$ & $35.6_{0.0}$ & $38.8_{0.1}$ & $62.3_{0.3}$ & $53.9_{0.1}$ & $57.8_{0.1}$ \\
	Description & $2.3_{0.1}$ & $4.4_{0.1}$ & $3.1_{0.1}$ & $6.4_{0.1}$ & $6.5_{0.1}$ & $6.4_{0.1}$ & $44.9_{0.1}$ & $43.8_{0.4}$ & $44.4_{0.1}$ & $64.1_{0.1}$ & $63.2_{0.1}$ & $63.6_{0.1}$ \\
	Importance & $2.6_{0.2}$ & $3.8_{0.3}$ & $3.1_{0.2}$ & $5.9_{0.0}$ & $7.2_{0.1}$ & $6.5_{0.0}$ & $39.5_{0.2}$ & $39.5_{0.3}$ & $39.5_{0.1}$ & $50.6_{0.0}$ & $53.4_{0.1}$ & $52.0_{0.0}$ \\
    \midrule
	Macro Avg. & $7.5_{0.1}$ & $8.8_{0.0}$ & $7.9_{0.0}$ & $14.4_{0.0}$ & $15.4_{0.2}$ & $14.8_{0.1}$ & $42.4_{0.0}$ & $38.7_{0.1}$ & $40.4_{0.0}$ & $57.7_{0.0}$ & $55.3_{0.1}$ & $56.4_{0.0}$ \\
    \bottomrule
    \end{tabular}}
    \caption{
        Main results of fine-grained entity typing~(ET).
        We report precision, recall, and F1 scores based on exact matching of type names in this table.
        \DevEx
    }
    \label{tab:main-results-typing}
\end{table*}

\begin{table*}[t]
    \centering
    \small
    \setlength{\tabcolsep}{0.5mm}{
    \begin{tabular}{ll|cccc}
    \toprule
    Partition & Instruction & ChatGPT & ChatGPT w/Demo & \modelname-1b & \modelname-7b\\
    \midrule
	 \multirow{6}{*}{\makecell[c]{Before\\05/30/2022}} & Default & $4.0_{0.0}$ & $7.8_{0.1}$ & $41.9_{0.1}$ & $63.4_{0.0}$ \\
	 & Base Type & $21.1_{0.4}$ & $34.4_{0.7}$ & $46.2_{0.2}$ & $71.3_{0.1}$ \\
	 & Abstract Type & $8.4_{0.2}$ & $19.2_{0.1}$ & $40.0_{0.0}$ & $65.6_{0.3}$ \\
	 & Description & $5.0_{0.1}$ & $7.6_{0.1}$ & $50.6_{0.5}$ & $73.3_{0.1}$ \\
	 & Importance & $4.2_{0.5}$ & $7.6_{0.3}$ & $47.1_{0.3}$ & $65.5_{0.1}$ \\
	\cline{2-6}
	& Macro Avg. & $9.7_{0.1}$ & $15.9_{0.2}$ & $44.4_{0.1}$ & $66.8_{0.1}$ \\
	\midrule
	 \multirow{6}{*}{\makecell[c]{After\\05/30/2022}} & Default & $2.6_{0.0}$ & $5.9_{0.3}$ & $17.7_{0.1}$ & $13.8_{0.0}$ \\
	 & Base Type & $16.1_{0.6}$ & $36.2_{0.0}$ & $27.2_{0.3}$ & $30.2_{0.0}$ \\
	 & Abstract Type & $5.8_{0.5}$ & $16.3_{0.3}$ & $27.4_{0.1}$ & $32.2_{0.3}$ \\
	 & Description & $2.7_{0.4}$ & $3.1_{0.2}$ & $21.0_{0.2}$ & $29.1_{0.1}$ \\
	 & Importance & $2.7_{0.4}$ & $6.0_{0.9}$ & $12.7_{0.1}$ & $10.7_{0.2}$ \\
	\cline{2-6}
	& Macro Avg. & $6.9_{0.1}$ & $14.2_{0.1}$ & $22.8_{0.0}$ & $24.1_{0.2}$ \\
    \bottomrule
    \end{tabular}}
    \caption{
        Results of generalization study on entity typing~(ET).
        We report recalls of entity types based on the exact matching of type names.
        \PartitionEx
        \DevEx
    }
    \label{tab:main-results-typing-generalization}
\end{table*}

\begin{table*}[t]
    \centering
    \small
    \setlength{\tabcolsep}{1mm}{
    \begin{tabular}{ll|cccc}
    \toprule
    Partition & Instruction & ChatGPT & ChatGPT w/Demo & \modelname-1b & \modelname-7b\\
    \midrule
	 \multirow{6}{*}{\makecell[c]{Before\\05/30/2022}} & Default & $36.2_{0.2}$ & $46.3_{0.1}$ & $74.3_{0.0}$ & $83.4_{0.0}$ \\
	 & Base Type & $38.9_{0.6}$ & $48.7_{0.2}$ & $67.4_{0.1}$ & $79.5_{0.1}$ \\
	 & Abstract Type & $40.6_{1.5}$ & $56.2_{1.4}$ & $67.5_{0.4}$ & $81.3_{0.1}$ \\
	 & Description & $57.8_{0.5}$ & $85.9_{0.6}$ & $75.8_{0.3}$ & $86.8_{0.1}$ \\
	 & Importance & $38.2_{0.5}$ & $46.5_{0.1}$ & $75.9_{0.2}$ & $84.6_{0.1}$ \\
	\cline{2-6}
	& Macro Avg. & $40.2_{0.4}$ & $53.6_{0.2}$ & $71.4_{0.1}$ & $82.5_{0.0}$ \\
	\midrule
	 \multirow{6}{*}{\makecell[c]{After\\05/30/2022}} & Default & $36.8_{0.3}$ & $45.0_{0.2}$ & $51.6_{0.1}$ & $60.7_{0.1}$ \\
	 & Base Type & $37.9_{0.9}$ & $44.6_{1.1}$ & $54.7_{0.4}$ & $60.0_{0.3}$ \\
	 & Abstract Type & $28.7_{0.3}$ & $44.1_{0.8}$ & $51.3_{0.4}$ & $58.8_{0.0}$ \\
	 & Description & $47.7_{0.7}$ & $75.9_{0.6}$ & $81.5_{0.6}$ & $93.1_{0.2}$ \\
	 & Importance & $36.0_{1.1}$ & $44.1_{0.8}$ & $45.0_{0.8}$ & $57.9_{0.3}$ \\
	\cline{2-6}
	& Macro Avg. & $36.0_{0.3}$ & $48.1_{0.2}$ & $55.7_{0.3}$ & $64.1_{0.2}$ \\
    \bottomrule
    \end{tabular}}
    \caption{
        Results of generalization study on entity description generation.
        We report average ROUGE-L F1 scores in this table.
        \PartitionEx
        \DevEx
    }
    \label{tab:main-results-desc}
\end{table*}

\begin{table*}[t]
    \centering
    \small
    \setlength{\tabcolsep}{1mm}{
    \begin{tabular}{ll|cccc}
    \toprule
    Partition & Instruction & ChatGPT & ChatGPT w/Demo & \modelname-1b & \modelname-7b\\
    \midrule
	\multirow{6}{*}{\makecell[c]{Before\\05/30/2022}} & Default & $8.5_{0.1}$ & $1.1_{0.0}$ & $69.8_{0.0}$ & $78.0_{0.0}$ \\
	 & Base Type & $9.1_{0.4}$ & $7.7_{0.2}$ & $71.3_{0.1}$ & $76.7_{0.1}$ \\
	 & Abstract Type & $10.1_{0.4}$ & $14.8_{0.2}$ & $67.3_{0.4}$ & $78.7_{0.2}$ \\
	 & Description & $10.1_{0.4}$ & $19.4_{0.5}$ & $65.6_{0.2}$ & $76.2_{0.1}$ \\
	 & Importance & $10.1_{0.6}$ & $1.1_{0.3}$ & $75.1_{0.1}$ & $81.6_{0.1}$ \\
	\cline{2-6}
	& Macro Avg. & $10.1_{0.1}$ & $8.4_{0.0}$ & $69.7_{0.1}$ & $78.0_{0.0}$ \\
	\midrule
	 \multirow{6}{*}{\makecell[c]{After\\05/30/2022}} & Default & $10.9_{0.6}$ & $1.8_{0.1}$ & $16.9_{0.1}$ & $21.1_{0.1}$ \\
	 & Base Type & $14.6_{1.3}$ & $11.9_{1.8}$ & $22.7_{0.3}$ & $24.5_{0.5}$ \\
	 & Abstract Type & $17.1_{2.5}$ & $21.4_{1.3}$ & $23.6_{0.2}$ & $27.5_{0.3}$ \\
	 & Description & $15.7_{1.6}$ & $37.4_{1.2}$ & $18.9_{0.1}$ & $17.9_{0.3}$ \\
	 & Importance & $14.0_{1.4}$ & $1.8_{0.2}$ & $24.5_{0.2}$ & $28.7_{0.2}$ \\
	\cline{2-6}
	& Macro Avg. & $14.3_{0.3}$ & $13.3_{0.1}$ & $20.5_{0.2}$ & $22.7_{0.1}$ \\
    \bottomrule
    \end{tabular}}
    \caption{
    Results of generalization study on entity aliases generation.
    We report recalls of entity aliases based on exact matching in this table.
    \PartitionEx
    \DevEx
    }
    \label{tab:main-results-aliases}
\end{table*}

\begin{table*}[t]
    \centering
    \small
    \setlength{\tabcolsep}{0.3mm}{
    \begin{tabular}{l|ccc|ccc|ccc|ccc}
    \toprule
    \multirow{2}{*}{Instruction} & \multicolumn{3}{c}{ChatGPT}\vline & \multicolumn{3}{c}{ChatGPT w/Demo}\vline & \multicolumn{3}{c}{\modelname-1b}\vline & \multicolumn{3}{c}{\modelname-7b} \\
     & MD & IE(T=1) & IE(T=0.8) & MD & IE(T=1) & IE(T=0.8) & MD & IE(T=1) & IE(T=0.8) & MD & IE(T=1) & IE(T=0.8) \\
     \midrule
     Number & $51.4_{0.1}$ & $24.8_{0.6}$ & $25.9_{0.7}$ & $49.0_{0.1}$ & $37.5_{0.2}$ & $39.1_{0.2}$ & $69.9_{0.1}$ & $55.9_{0.1}$ & $57.1_{0.1}$ & $81.3_{0.1}$ & $71.3_{0.1}$ & $72.2_{0.1}$ \\
	\makecell[l]{Number+\\Base Type} & $53.1_{0.3}$ & $25.8_{0.3}$ & $27.1_{0.2}$ & $50.3_{0.0}$ & $38.4_{0.1}$ & $40.1_{0.0}$ & $69.4_{0.2}$ & $55.2_{0.2}$ & $56.9_{0.2}$ & $80.8_{0.1}$ & $71.2_{0.1}$ & $72.2_{0.1}$ \\
	\makecell[l]{Number+\\Abstract Type} & $53.3_{0.5}$ & $25.0_{0.2}$ & $26.2_{0.1}$ & $51.1_{0.1}$ & $38.0_{0.1}$ & $39.5_{0.1}$ & $72.4_{0.3}$ & $57.4_{0.2}$ & $59.0_{0.2}$ & $82.0_{0.1}$ & $73.6_{0.2}$ & $74.3_{0.2}$ \\
    \midrule
	Macro Avg. & $52.6_{0.3}$ & $25.2_{0.1}$ & $26.4_{0.2}$ & $50.1_{0.0}$ & $38.0_{0.1}$ & $39.6_{0.1}$ & $70.6_{0.1}$ & $56.2_{0.0}$ & $57.7_{0.0}$ & $81.4_{0.1}$ & $72.0_{0.1}$ & $72.9_{0.1}$ \\
    \bottomrule
    \end{tabular}}
    \caption{
    Results of partial extraction instructions on mention detection (MD) and information extraction (IE).
    We report precision for mention spans in MD and title matching with ROUGE-L F1 threshold 0.8 and 1 in IE.
    \PartitionEx
    \DevEx
    }
    \label{tab:main-results-partial-md-ie}
\end{table*}

\begin{table*}[t]
    \centering
    \small
    \setlength{\tabcolsep}{1mm}{
    \begin{tabularx}{\linewidth}{l
    >{\raggedright\arraybackslash}X
    }
    \toprule
    Category & Prompt \\
    \midrule
    Default & Context: ``Extract entities.''$\backslash$n$\backslash$n Please rephrase this context. \\
    Base Type & Context: ``Extract entities in types \{types\}.''$\backslash$n$\backslash$n \{types\} in the context is a placeholder for a list of entity types. Please rephrase this context and keep \{types\} in the rephrased sentence. \{types\} should be put after the word ``types''"\\
    Abstract Type & Context: ``Extract entities in types \{types\}.''$\backslash$n$\backslash$n \{types\} in the context is a placeholder for a list of entity types. Please rephrase this context and keep \{types\} in the rephrased sentence. \{types\} should be put after the word ``types''"\\
    Description & Context: ``Extract entities in following descriptions: \{descriptions\}''$\backslash$n$\backslash$n \{descriptions\} in the context is a placeholder for a list of entity descriptions. Please rephrase this context and keep \{descriptions\} in the rephrased sentence.\\
    Importance & Context: ``Extract the most important \{number\} entities.''$\backslash$n$\backslash$n \{number\} in the context is a placeholder for the number of entity. Please rephrase this context and keep \{number\} in the rephrased sentence.\\
    Number & Context: ``Extract \{number\} entities.''$\backslash$n$\backslash$n \{number\} in the context is a placeholder for the number of entity. Please rephrase this context and keep \{number\} in the rephrased sentence. \\
    Number+Base Type & Context: ``Extract \{number\} entities in types \{types\}.''$\backslash$n$\backslash$n \{types\} in the context is a placeholder for a list of entity types. \{number\} in the context is a placeholder for the number of entities. Please rephrase this context and keep \{types\} and \{number\} in the rephrased sentence. \{types\} should be put after the word ``types''" \\
    Number+Abstract Type & Context: ``Extract \{number\} entities in types \{types\}.''$\backslash$n$\backslash$n \{types\} in the context is a placeholder for a list of entity types. \{number\} in the context is a placeholder for the number of entities. Please rephrase this context and keep \{types\} and \{number\} in the rephrased sentence. \{types\} should be put after the word ``types''"\\
    \bottomrule
    \end{tabularx}}
    \caption{
        Details of ChatGPT prompts we used to rephrase manually designed templates.
    }
    \label{tab:rephrase-prompts}
\end{table*}

\begin{table*}[t]
    \centering
    \small
    \setlength{\tabcolsep}{1mm}{
    \begin{tabularx}{\linewidth}{l
    >{\raggedright\arraybackslash}X
    }
    \toprule
    Type & Prompt \\
    \midrule
    ChatGPT & [context]: \{context\}.$\backslash$n[instruction]: \{instruction\}.$\backslash$n$\backslash$nPlease provide the response in the JSON format. The response should contains entities and triplets. Each entity has its mention, title, a list of types, description, and a list of aliases. Each triplet has its head and tail mentions, and a list of relations. Here is an example of the return JSON format: \{"entities": [\{"mention": String, "title": String, "type": List[String], "description": String, "aliases":List[String]\}], "triplets": [\{"head": String, "tail": String, "relations": List[String]\}]\}. \\
    \bottomrule
    \end{tabularx}}
    \caption{
        Details of ChatGPT prompts we used to address instruction-following open-world IE.
        \{context\}, \{instruction\} are placeholders for input context, instruction, and one-shot example, respectively.
    }
    \label{tab:chatgpt-open-world-ie-prompts}
\end{table*}

\end{document}